\pdfoutput=1

\documentclass[11pt]{article}

\PassOptionsToPackage{dvipsnames}{xcolor}

\usepackage{emnlp2021}
\usepackage{times}
\usepackage{latexsym}
\usepackage[T1]{fontenc}
\usepackage[utf8]{inputenc}

\usepackage{hyperref}
\usepackage{graphicx}
\usepackage{subfig}
\usepackage{url}            %
\usepackage{amsthm}
\usepackage{amsmath}
\usepackage{amsfonts}       %
\usepackage{amssymb}
\usepackage{nicefrac}       %
\usepackage{algorithm}
\usepackage{algorithmic}
\usepackage{wrapfig}
\usepackage{graphicx}
\usepackage{enumitem}
\usepackage[symbol]{footmisc}
\usepackage{tabularx,colortbl}
\usepackage{multirow}
\usepackage{verbatim}
\usepackage[most]{tcolorbox}
\usepackage{pifont}
\newcommand{\cmark}{\ding{51}}%
\newcommand{\xmark}{\ding{55}}%
\usepackage{listings}
\usepackage{lstautogobble}  %
\usepackage{zi4}            %
\definecolor{bluekeywords}{rgb}{0.13, 0.13, 1}
\definecolor{greencomments}{rgb}{0, 0.5, 0}
\definecolor{redstrings}{rgb}{0.9, 0, 0}
\definecolor{graynumbers}{rgb}{0.5, 0.5, 0.5}
\definecolor{lightgray}{rgb}{.9,.9,.9}
\definecolor{darkgray}{rgb}{.4,.4,.4}
\definecolor{purple}{rgb}{0.65, 0.12, 0.82}
\lstdefinelanguage{JavaScript}{
  keywords={break, case, catch, continue, debugger, default, delete, do, else, false, finally, for, function, if, in, instanceof, new, null, return, switch, this, throw, true, try, typeof, var, void, while, with},
  morecomment=[l]{//},
  morecomment=[s]{/*}{*/},
  morestring=[b]',
  morestring=[b]",
  ndkeywords={class, export, boolean, throw, implements, import, this},
  keywordstyle=\color{blue}\bfseries,
  ndkeywordstyle=\color{darkgray}\bfseries,
  identifierstyle=\color{black},
  commentstyle=\color{purple}\ttfamily,
  stringstyle=\color{red}\ttfamily,
  sensitive=true
}

\makeatletter
\def\blfootnote{\gdef\@thefnmark{}\@footnotetext}
\makeatother

\newcommand{\thou}[0]{$\textsc{k}$}
\newcommand{\million}[0]{$\textsc{m}$}

\usepackage{array}
\newcolumntype{H}{>{\setbox0=\hbox\bgroup}c<{\egroup}@{}}
\definecolor{Gray}{HTML}{ABD5FF}  %
\newcommand{\codesnip}[1]{\texttt{\small #1}}

\newcommand{\ours}[0]{ContraCode}

\title{Contrastive Code Representation Learning}

\author{Paras Jain$^*$ \and Ajay Jain$^*$ \and Tianjun Zhang \and\\\textbf{Pieter Abbeel \and Joseph E. Gonzalez \and Ion Stoica} \\
	Department of EECS, UC Berkeley\\
  \texttt{\{parasj, ajayj, tianjunz,}\\\texttt{pabbeel, jegonzal, istoica\}@berkeley.edu}}

\begin{document}
\maketitle
\begin{abstract}
Recent work learns contextual representations of source code by reconstructing tokens from their context. For downstream semantic understanding tasks like code clone detection, these representations should ideally capture program functionality. However, we show that the popular reconstruction-based RoBERTa model is sensitive to source code edits, \textit{even when the edits preserve semantics}. We propose ContraCode: a contrastive pre-training task that learns code functionality, not form. ContraCode pre-trains a neural network to identify functionally similar variants of a program among many non-equivalent distractors. We scalably generate these variants using an automated source-to-source compiler as a form of data augmentation. Contrastive pre-training outperforms RoBERTa on an adversarial code clone detection benchmark by 39\% AUROC. Surprisingly, improved adversarial robustness translates to better accuracy over natural code; ContraCode improves summarization and TypeScript type inference accuracy by 2 to 13 percentage points over competitive baselines. All source is available at \url{https://github.com/parasj/contracode}.
\end{abstract}

\section{Introduction}
\footnotetext{$^*$ equal contribution}
Programmers increasingly rely on machine-aided programming tools that analyze or transform code automatically to aid software development~\citep{refactoring_kim2012field}. Traditionally, code analysis uses hand-written rules, though the wide diversity of programs encountered in practice can limit their generality. Recent work leverages machine learning for richer language understanding, such as learning to detect bugs \cite{pradel2018deepbugs} and predict performance \cite{mendis2019ithemal}.

\begin{figure}[t]
    \centering
    \includegraphics[trim={0 6mm 0 0},clip,width=\linewidth]{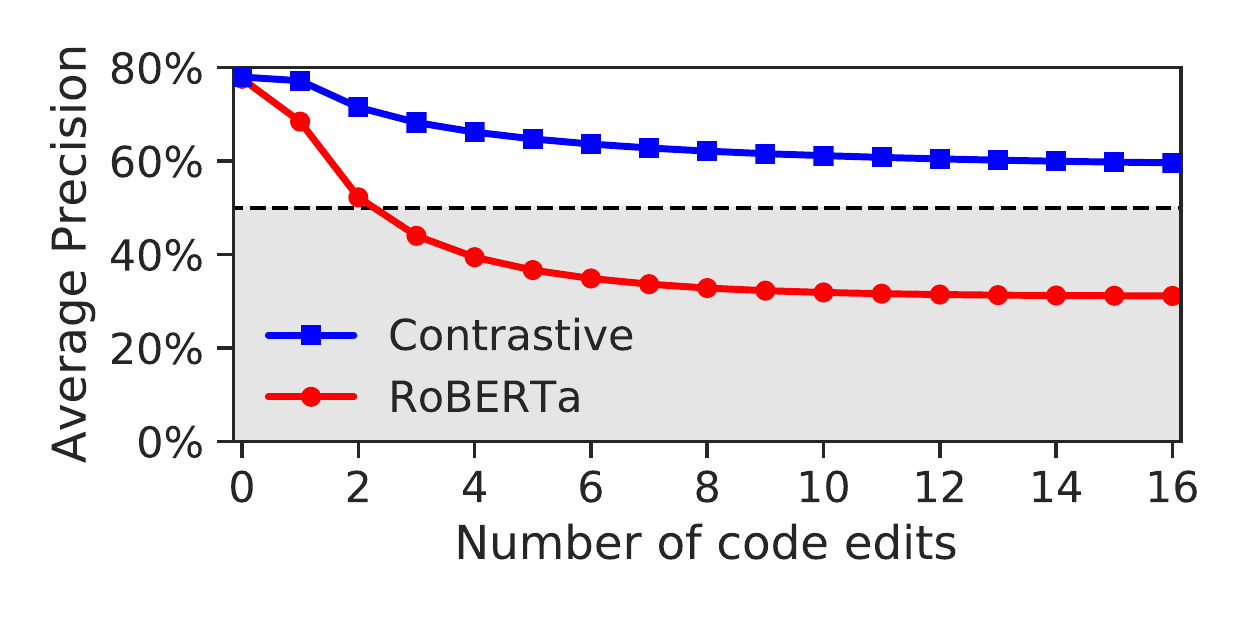}
    \caption{\textbf{Robust code clone detection:} On source code, \emph{RoBERTa is not robust to simple label-preserving code edits} like renaming variables. Adversarially selecting between possible edits lowers performance below random guessing (dashed line). Contrastive pre-training with \ours{} learns a more robust representation of functionality, consistent across code edits.}
    \label{fig:bert_motivation_robust}
\end{figure}

\begin{figure}[t]
    \centering
    \includegraphics[width=\linewidth]{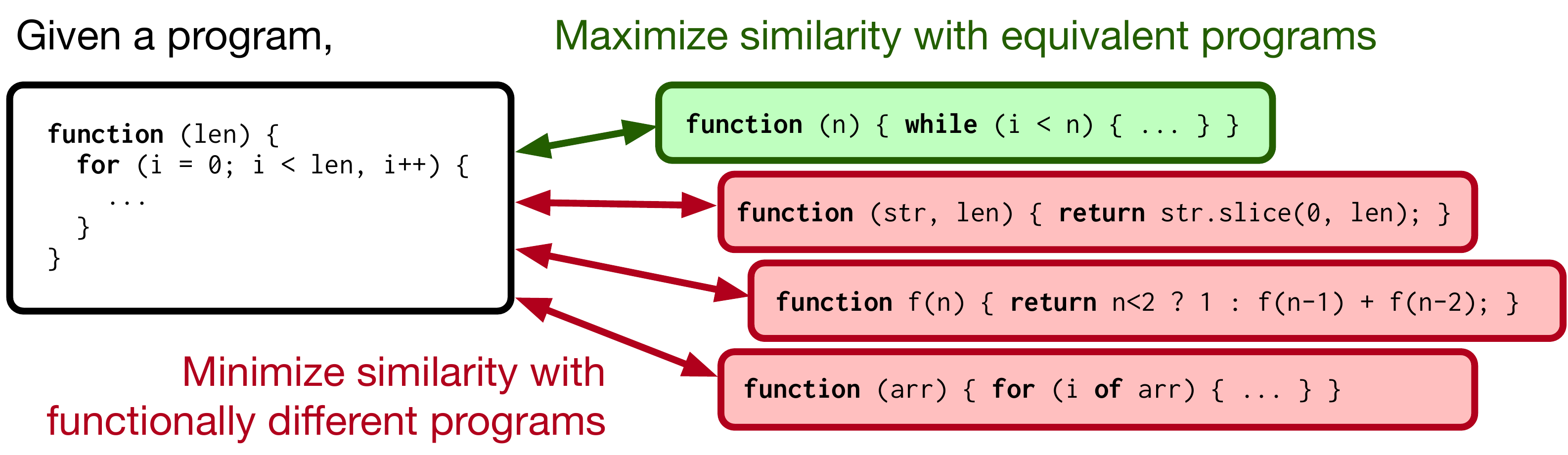}
    \caption{For many analyses, programs with the same functionality should have similar representations. \ours{} learns such representations by pre-training an encoder to retrieve equivalent, transformed programs among many distractors.}
    \label{fig:conceptual}
\end{figure}

Still, neural models of source code are susceptible to adversarial attacks. \citet{adversarial:yefet2020adversarial} and~\citet{adversarial:schuster2021you} find accuracy degrades significantly under adversarial perturbations for both discriminative and generative code models. In our work, we investigate adversarial attacks on code clone detection. Successful adversarial attacks could circumvent malware detectors.

\begin{figure}[t]
    \centering
    \includegraphics[width=\linewidth]{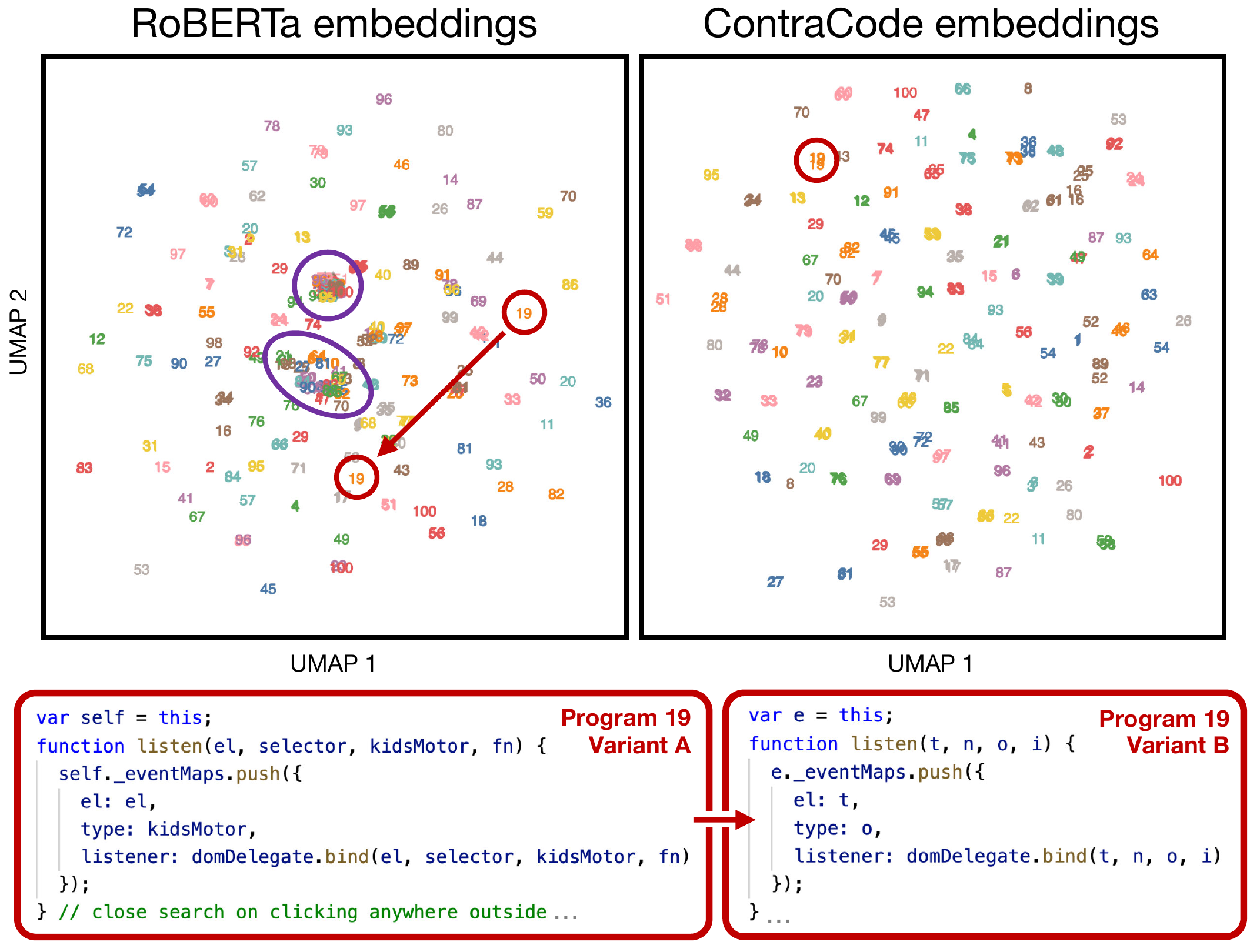}
    \caption{A UMAP visualization of JavaScript method representations learned by RoBERTa and \ours{}, in $\mathbb{R}^2$. Programs with the same functionality share color and number. RoBERTa's embeddings often do not cluster by functionality, suggesting that it is sensitive to implementation details. For example, many different programs {\color[HTML]{7030A0}\textbf{overlap}}, and renaming the variables of Program 19 significantly {\color[HTML]{C00000}\textbf{changes the embedding}}. In contrast, variants of Program 19 cluster in \ours{}'s embedding space.}
    \label{fig:bert_motivation_tsne}
\end{figure}

While self-supervision can improve adversarial robustness~\cite{hendrycks2019selfsupervised}, we find that RoBERTa is sensitive to stylistic implementation choices of code inputs. Fig.~\ref{fig:bert_motivation_robust} plots the performance of RoBERTa and ContraCode, our proposed method, on a code clone detection task as small label-preserving perturbations are applied to the input code syntax. With just three minor adversarial edits to code syntax, RoBERTa underperforms the random classifier (in gray). In Fig.~\ref{fig:bert_motivation_tsne}, we show that RoBERTa's representations of code are sensitive to code edits in agreement with prior work~\cite{wang2019coset, wang2019learning, rabin2020evaluation}.

To address this issue, we develop \ours{}: a self-supervised representation learning algorithm that captures program semantics. %
We hypothesize that \emph{programs with the same functionality should have similar underlying representations} for downstream code understanding tasks.

\ours{} generates syntactically diverse but functionally equivalent programs using source-to-source compiler transformation techniques (\textit{e.g.}, dead code elimination, obfuscation and constant folding).
It uses these programs in a challenging discriminative pretext task that requires the model to identify similar programs out of a large dataset of distractors (Fig.~\ref{fig:conceptual}). 
To solve this task, the model must embed code semantics rather than syntax. 
\ours{} improves adversarial robustness in Fig.~\ref{fig:bert_motivation_robust}. Surprisingly, adversarial robustness transfers to better natural code understanding.

Our novel contributions include:
\begin{enumerate}
    \item the novel use of compiler-based transformations as data augmentations for code,
    \item the concept of program representation learning based on functional equivalence, and
    \item a detailed analysis of architectures, code transforms and pre-training strategies, showing \ours{} improves type inference top-1 accuracy by 9\%, learned inference by 2\%--13\%, summarization F1 score by up to 8\% and clone detection AUROC by 2\%--46\%.
\end{enumerate}

\section{Related work}
\paragraph{Self-supervised learning} (SSL) is a learning strategy where some attributes of a datapoint are predicted from remaining parts.
BERT~\cite{devlin2018bert} is a SSL method for NLP that reconstructs masked tokens as a pretext task. RoBERTa~\cite{liu2019roberta} further tunes BERT. %
Contrastive approaches minimize distance between learned representations of similar examples (positives) and maximize distance between dissimilar negatives~\citep{hadsell2006dimensionality}.
CPC~\citep{cpcv1_oord2018representation, henaff2019data} encodes segments of sequential data to predict future segments. %
SimCLR~\citep{chen2020simple} and MoCo~\citep{he2019momentum, chen2020improved} use many negatives for dense loss signal.

\paragraph{Code representation learning}~~We address clone detection~\cite{white2016deep}, type inference~\citep{hellendoorn2018deep}, and summarization~\cite{alon2018code2seq}.
Others explored summarization~\citep{movshovitz2013natural, allamanis2016convolutional, iyer2016summarizing, ahmad2020summarization} and types~\citep{pradel2019typewriter, p2020opttyper, Wei2020LambdaNet, allamanis2020typilus, bielik2020adversarial, allamanis2018survey} for various languages.
Inst2vec~\citep{ben2018neural} embeds statements in LLVM IR by processing a flow graph with a context prediction objective~\citep{mikolov2013distributed}.
Code2seq \citep{alon2018code2seq} embeds AST paths with an attentional encoder for seq2seq tasks.
\citet{cuBERT} and \citet{feng2020codebert} pre-train a Transformer on code using the masked language modeling (MLM) objective~\citep{devlin2018bert, taylor1953cloze}.

\begin{figure*}[t]
\centering
\begin{minipage}{.69\linewidth}
  \centering
  \includegraphics[width=\linewidth]{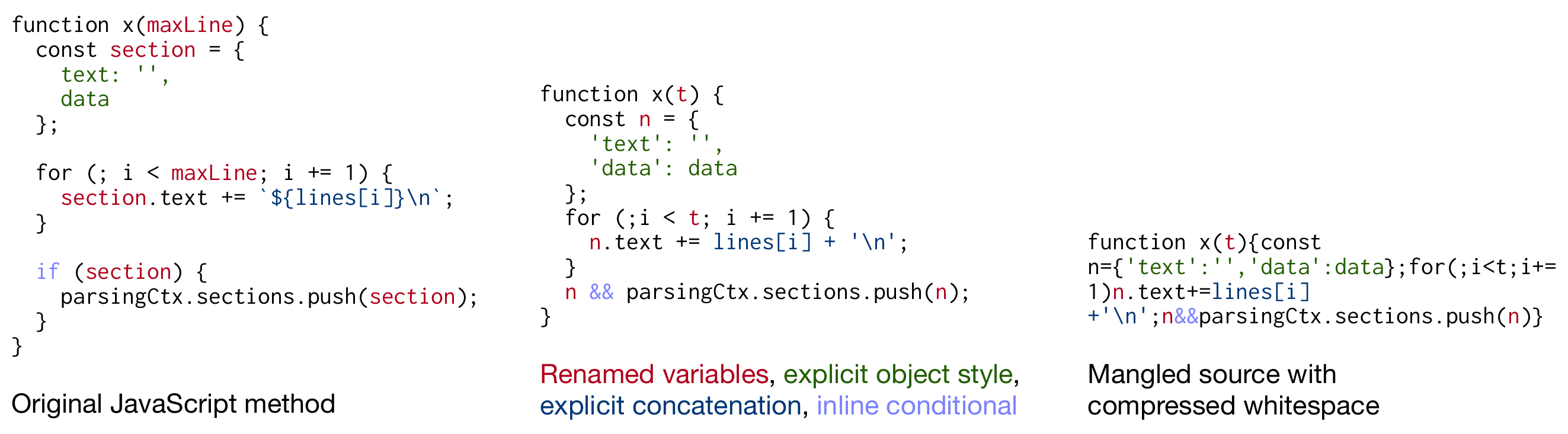}
  \captionof{figure}{A JavaScript method from our unlabeled training set with two automatically generated semantically-equivalent programs. The method is from the StackEdit Markdown editor.}
    \label{fig:augmentation_examples}
\end{minipage}\hfill%
\begin{minipage}{.29\linewidth}
  \centering
  \includegraphics[width=\linewidth]{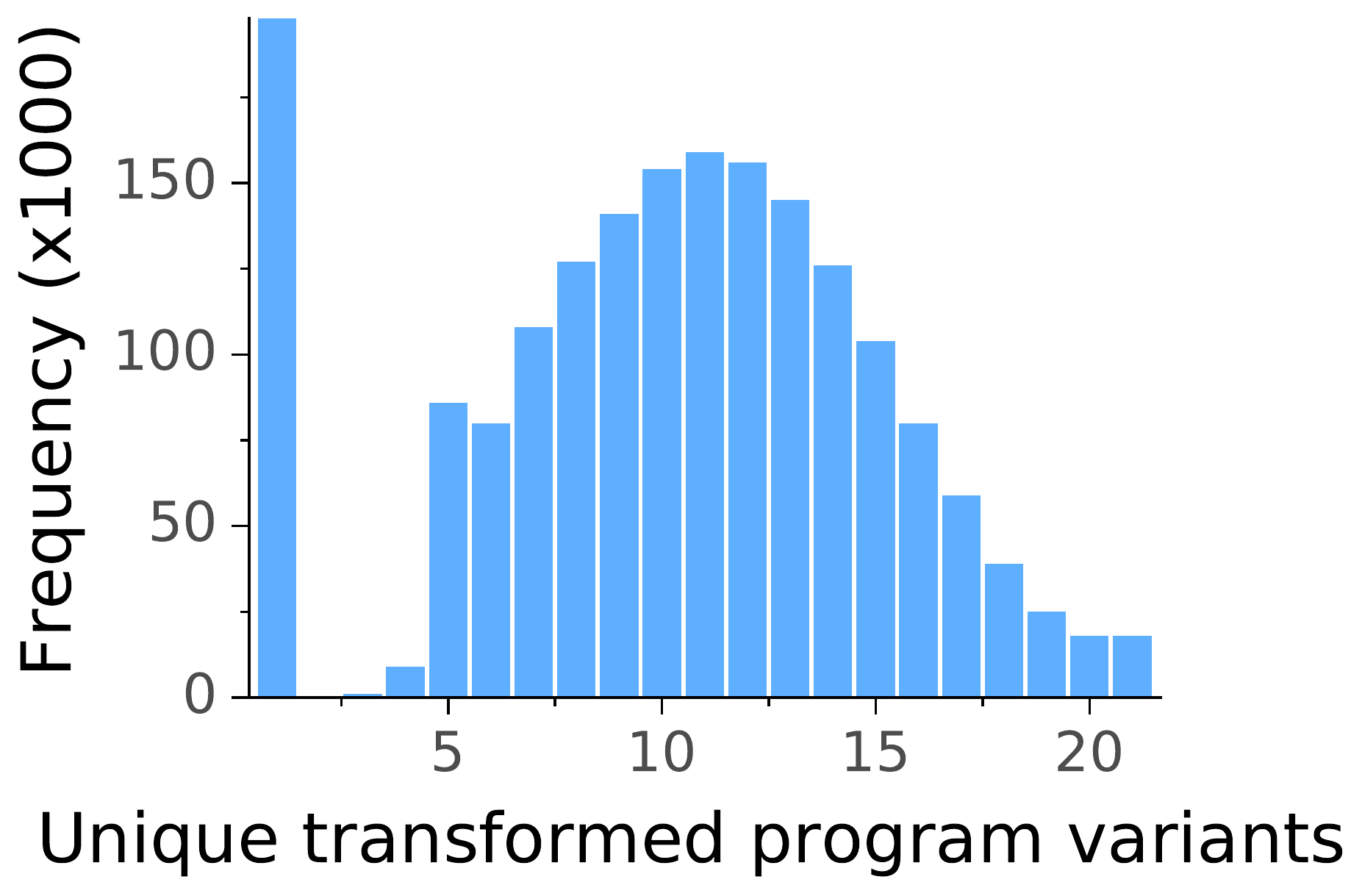}
  \captionof{figure}{Histogram of the number of unique transformed variants per JavaScript method during pre-training.}
  \label{fig:augmentation_histogram}
\end{minipage}
\end{figure*}

\paragraph{Adversarial attacks on code models}~~~\citet{yefet2019adversarial} find code models are highly sensitive to adversarial code edits in a discrimative setting. \citet{schuster1997bidirectional} discovers in-the-wild attacks on code autocompletion tools.
Compared to language models, code models may be more vulnerable to adversarial attacks due to synthetic labels~\citep{ferenc2018public,pradel2018deepbugs,benton2019defexts} and duplication~\citep{10.1145/3359591.3359735} that degrade generalization. %

\section{Approach}

Our core insight is to use compiler transforms as data augmentations, generating a dataset of equivalent functions (\S\ref{sec:data_aug}, \ref{sec:diverse_aug}). We then use a contrastive objective to learn a representation invariant to these transforms (\S\ref{infonce_loss}).

\subsection{Compilation as data augmentation} \label{sec:data_aug}

Modern programming languages afford great flexibility to software developers, allowing them to implement the same function in different ways. Yet, crowdsourcing equivalent programs from GitHub is difficult as verifying equivalence is undecidable~\citep{10.1145/512529.512566, 10.1145/1168857.1168906} and approximate verification is costly and runs untrusted code~\citep{massalin1987superoptimizer}.

Instead of searching for equivalences, we propose correct-by-construction data augmentation. We apply compiler transforms to unlabeled code to generates many variants with equivalent functionality, \textit{i.e.} operational semantics.
For example, dead-code elimination (DCE) is an optimization that removes operations that do not change function output. While DCE preserves functionality, \citet{wang2019coset} find that up to 12.7\% of the predictions of current supervised algorithm classification models change after DCE.

We parse a particular source code sequence, \textit{e.g.} \codesnip{W*x~+~b} into a tree-structured representation \codesnip{(+~(*~W~x)~b)} called an Abstract Syntax Tree (AST).
We then transform the AST with automated traversal passes.
A rich body of prior programming language work explores parsing and transforming ASTs to optimize a program. If source code is emitted by the compiler rather than machine code, this is called source-to-source transformation or transpilation. Transpilation is common for optimizing and obfuscating dynamic languages like JavaScript. Further, if each transform preserves code semantics, then any composition also preserves semantics.

\begin{table}[t]
	\resizebox{\linewidth}{!}{
    \begin{tabular}{@{}llll@{}}
    \hline
           & \textbf{\large Code compression}          &        & \textbf{\large Identifier modification}           \\
    \cmark & Reformatting (R)                   & \cmark &  Variable renaming (VR)             \\
    \cmark & Beautification (B)                 & \cmark &  Identifier mangling (IM)           \\
    \cmark & Compression (C)                    &        & \textbf{\large Regularization}             \\ 
    \cmark & Dead-code elimination (DCE)        & \cmark & Dead-code insertion (DCI)           \\
    \cmark & Type upconversion (T)              & \cmark & Subword regularization (SW)         \\
    \cmark & Constant folding (CF)              & \xmark & Line subsampling (LS)               \\ \hline \vspace{-3mm} \\
    \multicolumn{4}{l}{\cmark~= semantics-preserving transformation~~~~\xmark~= lossy transformation} \\
    \end{tabular}
    }
    \caption{We augment programs with 11 automated source-to-source compiler transforms. 10 are correct-by-construction and preserve operational semantics.}%
    \label{tab:list_of_transformations}
\end{table}

We implement our transpiler with the Babel and Terser compiler infrastructures~\citep{babel_github,terser_github} for the JavaScript programming language. In future work, a language-agnostic compiler~\cite{10.1145/3276492} could be used to extend \ours{} to other languages. Each compiler transformation is a function ${\tau : \mathcal{P} \rightarrow \mathcal{P}}$, where the space of programs $\mathcal{P}$ is composed of the set of valid ASTs and the set of programs in tokenized source form. Fig.~\ref{fig:augmentation_examples} shows variants of an example program.
Table~\ref{tab:list_of_transformations} and Appendix~\ref{sec:appendix:program_transformations} list program transformations in detail, but we broadly group them into three categories:
\begin{itemize}
    \item \textbf{Code compression} changes the syntactic structure of code and performs correct-by-construction transforms such as pre-computing constant expressions.
    \item \textbf{Identifier modifications} substitute method and variable names with random tokens, masking some human-readable information in a program but preserving functionality.
    \item Finally, \textbf{Regularizing transforms} improve model generalization by reducing the number of trivial positive pairs with high text overlap. The line subsampling pass in this group potentially modifies program semantics.
\end{itemize}

\subsection{Diversity through transform dropout} \label{sec:diverse_aug}

Stochastic augmentations in other modalities like random crops generate diverse outputs, but most of our compiler-based transformations are deterministic. To produce a diverse set of transformed programs, we randomly apply a subset of available compiler passes in a pre-specified order, applying transform $\tau_i$ with probability $p_i$. Intermediate programs are converted between AST and source form as needed for the compiler. Algorithm~\ref{alg:transformation} details our transform dropout procedure.

\begin{figure*}[t]
    \centering
    \includegraphics[width=\linewidth]{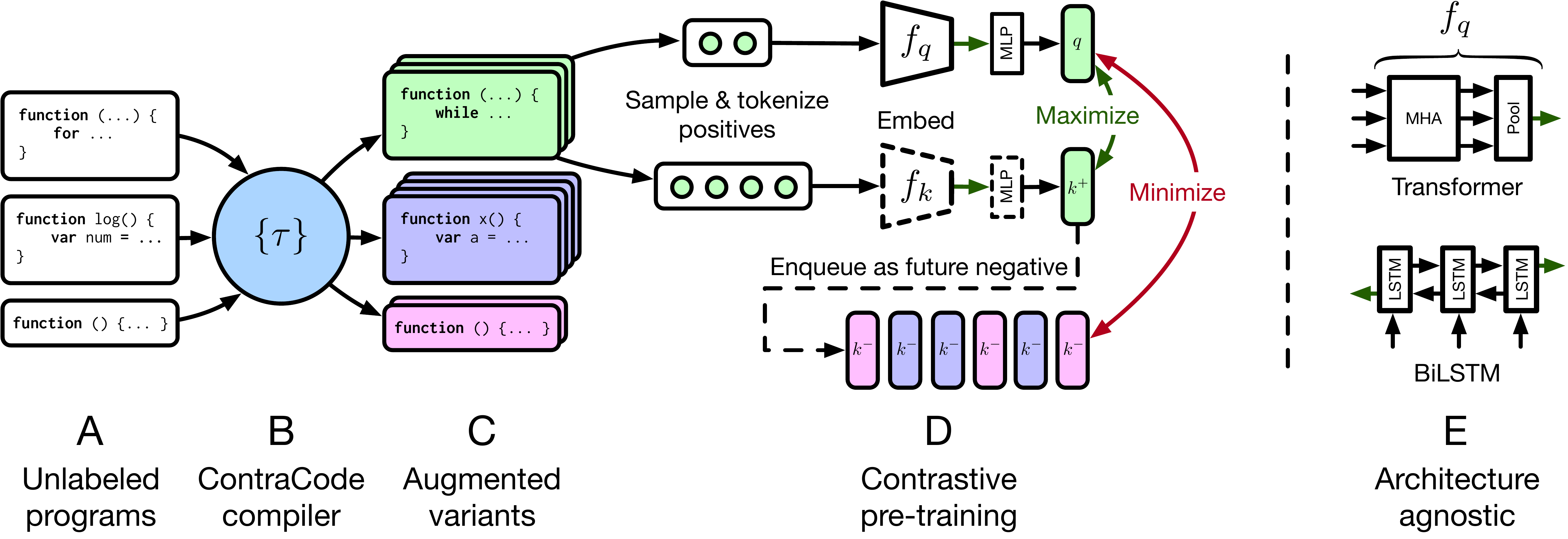}
    \caption{\ours{} pre-trains a neural program encoder $f_q$ and transfers it to downstream tasks. \textbf{A-B.} Unlabeled programs are transformed \textbf{C.} into augmented variants. \textbf{D.} We pre-train $f_q$ by maximizing similarity of projected embeddings of \textit{positive} program pairs--variants of the same program--and minimizing similarity with a queue of cached negatives. \textbf{E.} \ours{} supports any architecture for $f_q$ that produces a global program embedding such as Transformers and LSTMs. $f_q$ is then fine-tuned on smaller labeled datasets.}
    \label{fig:training}
\end{figure*}

Figure~\ref{fig:augmentation_histogram} measures the resulting diversity in programs. We precompute up to 20 augmentations of 1.8\million{} JavaScript methods from GitHub. Algorithm~\ref{alg:transformation} deduplicates method variants before pre-training since some transforms will leave the program unchanged. 89\% of the methods have more than one alternative after applying 20 random sequences of transformations.
The remaining methods without syntactically distinct alternatives include one-line functions that are obfuscated.
We apply subword regularization~\citep{kudo2018subword} as a final transformation to derive different tokenizations every batch, so pairs derived from the same original method will still differ.
All transformations are fast; our compiler transforms 300 functions per second on a single CPU core.

\begin{algorithm}[t]
\small
   \caption{\textbf{Transform dropout} for stochastic program augmentation.}
   \label{alg:transformation}
\begin{algorithmic}[1]
   \STATE {\bfseries Input:} Program source $x$, transformation functions $\tau_1,\ldots\tau_k$, transform probabilities $p_1,\ldots p_k$, count $N$
   \STATE {\bfseries Returns:} $N$ variants of $x$
   \STATE $\mathcal{V} \leftarrow \{x\}$, a set of augmented program variants
   \FOR{\textsc{Sample} $i \leftarrow 1 \ldots N-1$}
        \STATE $x' \leftarrow x$
   		\FOR{transform $t \leftarrow 1 \ldots k$}
   		    \STATE Sample $y_t \sim \text{Bernoulli}(p_t)$
   		    \IF{$y_t = 1$}
       		    \STATE \textbf{if} \textsc{RequiresAST}$(\tau_t(\cdot))$ and $\neg$\textsc{IsAST}$(x')$ \textbf{then} $x' \leftarrow \textsc{ParseToAST}(x')$
       		    \STATE \textbf{else if} $\neg$\textsc{RequiresAST}$(\tau_t(\cdot))$ and \textsc{IsAST}$(x')$ \textbf{then} $x' \leftarrow \textsc{LowerToSource}(x')$
    	        \STATE $x' \leftarrow \tau_t(x')$
    	    \ENDIF{}
   		\ENDFOR{}
   		\STATE \textbf{if} $\textsc{IsAST}(x')$ \textbf{then} $x' \leftarrow \textsc{LowerToSource}(x')$
   		\STATE $\mathcal{V} \leftarrow \mathcal{V} \cup \{x'\}$
   \ENDFOR{}
   \STATE {\bfseries return} $\mathcal{V}$
\end{algorithmic}
\end{algorithm}

\subsection{Contrastive pre-training} \label{infonce_loss}
We extend the Momentum Contrast (MoCo) methodology~\citep{he2019momentum} that was designed for contrastive image representation learning. In our case, we learn a program encoder $f_q$ that maps a sequence of program tokens to a single, fixed dimensional embedding.
We organize programs into \textit{functionally similar positive pairs} and \textit{dissimilar negative pairs}.
Generating two augmentations of the same GitHub program yields a positive pair $(x^q, x^{k^+})$, and an augmentation of a different program yields a negative $x^{k^-}$. The program $x^q$ is called a ``query'' used to retrieve the corresponding ``key'' $x^{k^+}$ during contrastive pre-training.
We use these to shape representation space, drawing positives together and pushing away from negatives. Negatives are important to prevent the encoder $f_q$ from mapping all programs to the same, trivial representation~\cite{pmlr-v97-saunshi19a}.

\paragraph{Pre-training objective}
Like \citet{he2019momentum}, we use the InfoNCE loss~\citep{cpcv1_oord2018representation}, a tractable objective that frames contrastive learning as a classification task: can the positives be identified among negatives? InfoNCE computes the probability of selecting the positive by taking the softmax of projected embedding similarities across a batch and a queue of negatives.
Eq. \eqref{eq:constrastive_loss} shows the InfoNCE loss, a function whose value is low when $q$ is similar to the positive key embedding $k^{+}$ and dissimilar to negative key embeddings $k^{-}$. $t$ is a temperature hyperparameter proposed by \citet{wu2018unsupervised}.
\begin{equation}\label{eq:constrastive_loss}
-\log \frac{\exp(q \cdot k^{+}/t)}{\exp(q \cdot k^{+}/t) + \sum_{k^{-}} \exp(q \cdot k^{-}/t)}
\end{equation}
The query representation $q = f_{q}(x^{q})$ is computed by the encoder network $f_q$, and $x^{q}$ is a query program. Likewise, $k = f_{k}(x^{k})$ using a separate key encoder $f_{k}$.
The summation $\sum_{k^-}$ in the normalizing denominator is taken over the queue of pre-computed negatives in the batch.

Following \citet{he2019momentum}, to reduce memory consumption during pre-training, we cache embedded programs from past batches in a queue containing negative samples, as shown in Fig.~\ref{fig:training}.
The query encoder $f_q$ is trained via gradient descent while the key encoder $f_k$ is trained slowly via an exponential moving average (EMA) of the query encoder parameters. The EMA update stabilizes the pre-computed key embeddings across training iterations. Since keys are only embedded once per epoch, we use a very large set of negatives, over $100\textsc{K}$, with minimal additional computational cost and no explicit hard negative mining.

\ours{} is agnostic to the architecture of the program encoder $f_q$. We evaluate contrastive pre-training of 6-layer Transformer~\citep{vaswani2017attention} and 2-layer BiLSTM~\citep{schuster1997bidirectional, huang2015bidirectional} architectures (\S\ref{sec:experiments}).

\paragraph{Transfer learning} After pre-training converges, the encoder $f_q$ is transferred to downstream tasks. For code clone detection, we use $f_q(x)$ without fine-tuning. For tasks where the output space differs from the encoder, we add a task-specific MLP or Transformer decoder after $f_q$, then fine-tune the resulting network end-to-end on labeled task data.

\section{Evaluation}
\label{sec:experiments}
In order to evaluate whether \ours{} defend against adversarial code inputs, we benchmark adversarial code clone detection accuracy~\citep{Baker92aprogram}. We evaluate results over natural and adversarial edits. We then evaluate how improvements to adversarial robustness translate to improvements on established in-the-wild code benchmarks. While improvements on adversarial benchmarks would not be expected to translate to real code, we find significant improvements in extreme code summarization~\citep{allamanis2016convolutional} and type inference~\citep{hellendoorn2018deep} tasks.

Clone detection experiments show that contrastive and hybrid representations with our compiler-based augmentations are predictive of program functionality in-the-wild, and that contrastive representations are the most robust to adversarial edits (\S\ref{sec:experiments_code_clone}).
Contrastive pre-training outperforms baseline supervised and self-supervised methods on all three tasks (\S\ref{sec:experiments_code_clone}-\ref{sec:experiments_code_summarization}). Finally, ablations suggest it is better to augment unlabeled programs during pre-training rather than augmenting smaller supervised datasets (\S\ref{sec:experiments_augmentation}).

\paragraph{Experimental setup}
Models are pre-trained on CodeSearchNet, a large corpus of methods extracted from popular GitHub repositories~\citep{husain2019codesearchnet}. CodeSearchNet contains 1,843,099 JavaScript programs. Only 81,487 methods have both a documentation string and a method name. The asymmetry between labeled and unlabeled programs stems from JavaScript coding practices where anonymous functions are widespread. The pre-training dataset described in Section~\ref{sec:data_aug} is the result of augmenting all 1.8\million{} programs.

\begin{table*}
\setlength\tabcolsep{4.5pt}
\centering
\begin{tabular}{lccccccHH} \hline
 & \multicolumn{2}{c}{\textbf{Natural code}} & \multicolumn{2}{c}{\textbf{Adversarial} ($N$=4)} & \multicolumn{2}{c}{\textbf{Adversarial} ($N$=16)} \\
 & \textbf{AUROC} & \textbf{AP} & \textbf{AUROC} & \textbf{AP} & \textbf{AUROC} & \textbf{AP} & \textbf{AUROC} & \textbf{AP} \\ \hline
Edit distance heuristic   & 69.55\tiny{$\pm$0.81} & 73.75 & 31.63\tiny{$\pm$0.82} & 42.85 & 12.11\tiny{$\pm$0.54} & 32.46 \\
Randomly initialized Transformer & 72.31\tiny{$\pm$0.79} & 75.82 & 22.72\tiny{$\pm$0.20} & 37.73 & 3.09\tiny{$\pm$0.28} & 30.95 & 16.53 & 36.66 \\
~~~~+ RoBERTa MLM pre-train & 74.04\tiny{$\pm$0.77} & 77.65 & 25.83\tiny{$\pm$0.21} & 39.46 & 4.51\tiny{$\pm$0.33} & 31.17 & 18.78 & 37.57 \\
~~~~+ \cellcolor{Gray}ContraCode pre-train & \cellcolor{Gray}75.73\tiny{$\pm$0.75} & \cellcolor{Gray}78.02 & \cellcolor{Gray}\textbf{64.97}\tiny{$\pm$0.24} & \cellcolor{Gray}\textbf{66.23} & \cellcolor{Gray}\textbf{58.32}\tiny{$\pm$0.88} & \cellcolor{Gray}\textbf{59.66} & \textbf{62.72} & \textbf{64.06} \\
~~~~+ \cellcolor{Gray}ContraCode + RoBERTa MLM & \cellcolor{Gray}\textbf{79.39}\tiny{$\pm$0.70} & \cellcolor{Gray}\textbf{81.47} & \cellcolor{Gray}37.81\tiny{$\pm$0.24} & \cellcolor{Gray}51.42 & \cellcolor{Gray}10.09\tiny{$\pm$0.50} & \cellcolor{Gray}32.52 & 27.52 & 44.19\\ \hline
\end{tabular}
\caption{\textbf{Zero-shot code clone detection} with cosine similarity probe. Contrastive and hybrid representations improve clone detection AUROC on unmodified (natural) HackerRank programs by $+8\%$ and $+10\%$ AUROC over a heuristic textual similarity probe, respectively, suggesting they are predictive of functionality. Contrastive representations are also the most robust to adversarial code transformations.}
\label{tab:code_clone}
\end{table*}

\begin{figure*}[t]
    \centering
    \includegraphics[width=\linewidth]{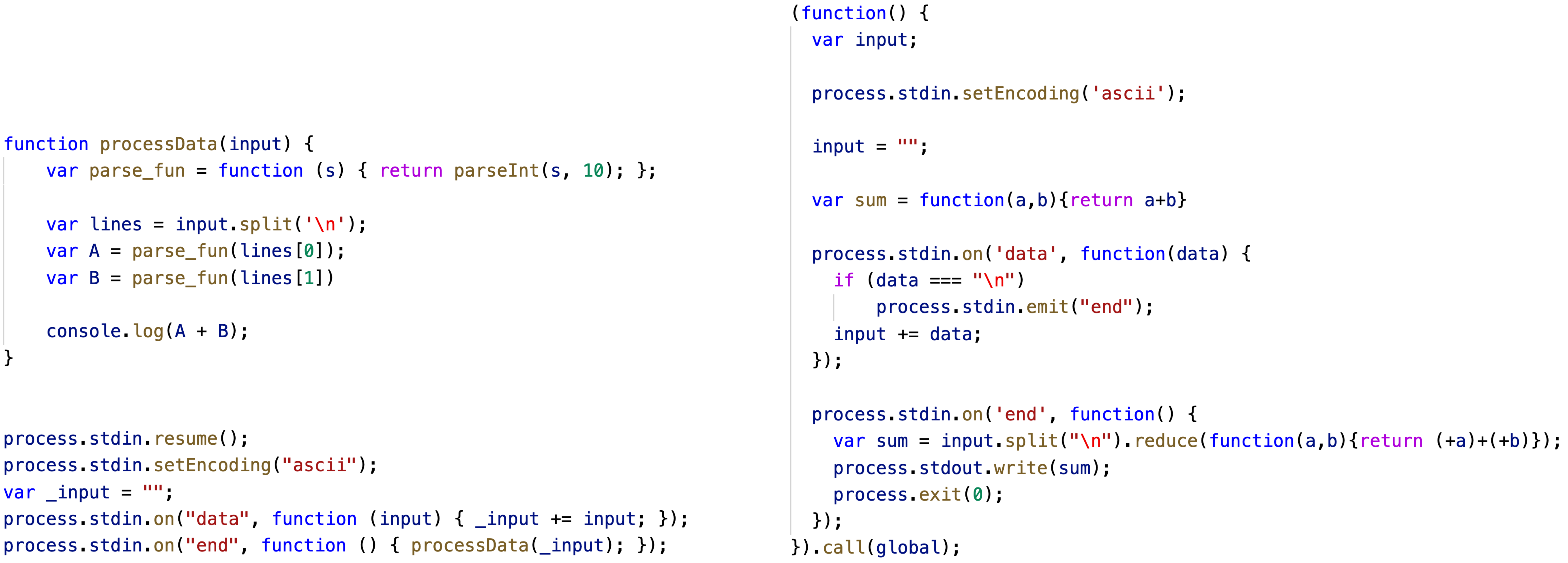}
    \caption{Code clone detection example. These programs solve the same HackerRank coding challenge (reading and summing two integers), but use different coding conventions. The neural code clone detector should classify this pair as a positive, \textit{i.e.} a clone.}
    \label{fig:code_clone_example}
\end{figure*}

As our approach supports any encoder, we evaluate two architectures: a 2-layer Bidirectional LSTM with 18\million{} parameters, similar to the supervised model used by \citet{hellendoorn2018deep}, and a 6-layer Transformer with 23\million{} parameters.
For a baseline self-supervised approach, we pre-train both architectures with the RoBERTa MLM objective, then transfer it to downstream tasks.

\subsection{Robust Zero-shot Code Clone Detection}
\label{sec:experiments_code_clone}

\ours{} learns to match variants of programs with similar functionality. While transformations produce highly diverse token sequences~(quantified in the supplement), they are artificial and do not change the underlying algorithm. In contrast, human programmers can solve a problem with many data structures, algorithms and programming models. To determine whether pre-trained representations are consistent across programs written by different people, we benchmark \textit{code clone detection}, a binary classification task to detect whether two programs solve the same problem or different ones (Fig.~\ref{fig:code_clone_example}). This is useful for deduplicating, refactoring and retrieving code, as well as checking approximate code correctness.

Benchmarks exist like BigCloneBench~\citep{10.1109/ICSME.2014.77}, but to the best of our knowledge, there is no benchmark for the JavaScript. We collected 274 in-the-wild JavaScript programs that correctly solve 33 problems from the HackerRank interview preparation website. 
There are 2065 pairs solving the same problem and 70\thou{} pairs solving different problems, which we randomly subsample to 2065 to balance the classes.

Since we probe zero-shot performance based on pre-trained representations, there is no training set. Instead, we threshold cosine similarity of pooled representations of the programs $u$ and $v$: $u^T v / \|u\|\|v\|$.
Many code analysis methods for clone detection measure textual similarity~\cite{Baker92aprogram}. As a baseline, we threshold the dissimilarity score, a scaled Levenshtein edit distance between normalized and tokenized programs.

Table~\ref{tab:code_clone} reports the area under the ROC curve (AUROC) and average precision (AP, area under Precision-Recall). All learned representations improve over the heuristic on natural code. Self-supervision through RoBERTa MLM pre-training improves over a randomly initialized network by +1.7\% AUROC. Contrastive pre-training achieves +3.4\% AUROC over the same baseline. A hybrid objective combining both the contrastive loss and MLM has the best performance with +7.0\% AUROC (+5.4\% over MLM alone). Although MLM is still useful over natural code, \ours{} learns overall stronger representations of functionality.

However, are these representations robust to code edits? We adversarially edit one program in each pair by applying the loss-maximizing code compression and identifier modification transformation among $N$ samples from Algorithm~\ref{alg:transformation}. These transformations preserve program functionality, so ground-truth labels are unchanged. With only 4 edits, RoBERTa performs worse than the heuristic (-5.8\% AUROC) and worse than random guessing (50\% AUROC), indicating it is highly sensitive to these kinds of implementation details. \ours{} retains much of its performance (+39\% AUROC over RoBERTa) as it explicitly optimizes for invariance to code edits. Surprisingly, the hybrid model is less robust than \ours{} alone, perhaps indicating that MLM learns non-robust features~\cite{NEURIPS2019_e2c420d9}.

\subsection{Fine-tuning for Type Inference}
\label{sec:experiments_type_inference}
JavaScript is a dynamically typed language, where variable types are determined at runtime based on the values they represent. Manually annotating code with types helps tools flag bugs by detecting incompatible types. Annotations also document code, but are tedious to maintain. Type inference tools automatically predict types from context.

To \textit{learn} to infer types, we use the annotated dataset of TypeScript programs from DeepTyper~\citep{hellendoorn2018deep}, excluding GitHub repositories that were made private or deleted since publication.
The training set contains 15,570 TypeScript files from 187 repositories with 6,902,642 total tokens. Validation and test sets are from held-out repositories.
For additional supervision, missing types are inferred by static analysis to augment user-defined types as targets.
A 2-layer MLP head predicts types from token embeddings output by the DeepTyper LSTM. We early stop based on validation set top-1 accuracy.

For the rest of our experiments, baseline RoBERTa models are pre-trained on the same \textit{augmented} data as \ours{} for fair comparison. Learning representations that transfer from unlabeled JavaScript programs is challenging because TypeScript supports a superset of JavaScript's grammar, with types annotations and other syntactic sugar that need to be learned during fine-tuning. Further, the pre-training data only has methods while DeepTyper's dataset uses entire files (modules). 
The model is only given source code for a single file, not dependencies.

\begin{table}[t]
\setlength\tabcolsep{1.2pt}
\centering
\begin{tabular}{lcc} \hline
    \textbf{Method}  & \textbf{Acc@1} & \textbf{Acc@5} \\ 
    \hline
    TypeScript CheckJS & 45.11\% & --- \\
    DeepTyper, variable name only & 28.94\% & 70.07\% \\ \hline
    GPT-3 Codex (zero-shot, 175B) & 26.62\% & --- \\
    GPT-3 Codex (few-shot, 175B) & 30.55\% & --- \\ \hline
    Transformer & 45.66\% & 80.08\% \\
    ~~+ RoBERTa MLM pre-train & 40.85\% & 75.76\%\\
    \cellcolor{Gray}~~+ \ours{} pre-train & \cellcolor{Gray}46.86\% & \cellcolor{Gray}\textbf{81.85}\%\\
    \cellcolor{Gray}~~+ \ours{} + MLM (hybrid) & \cellcolor{Gray}\textbf{47.16\%} & \cellcolor{Gray}81.44\%\\\hline
    DeepTyper BiLSTM & 51.73\% & 82.71\% \\
    ~~+ RoBERTa MLM pre-train & 50.24\% & 82.85\% \\
    \cellcolor{Gray}~~+ \ours{} pre-train & \cellcolor{Gray}\textbf{54.01\%} & \cellcolor{Gray}\textbf{85.55\%} \\
    \hline
\end{tabular}
\caption{\textbf{Type inference accuracy on TypeScript programs}. As \ours{} does not modify model architecture, contrastive pre-training improves both BiLSTM and Transformer accuracy (1.5\% to 2.28\%). Compared with TypeScript's built-in type inference, we improve accuracy by 8.9\%.}
\label{tab:types}
\end{table}

\begin{figure*}[t]
    \centering
    \includegraphics[trim={6mm 0 6mm 0},clip,width=\linewidth]{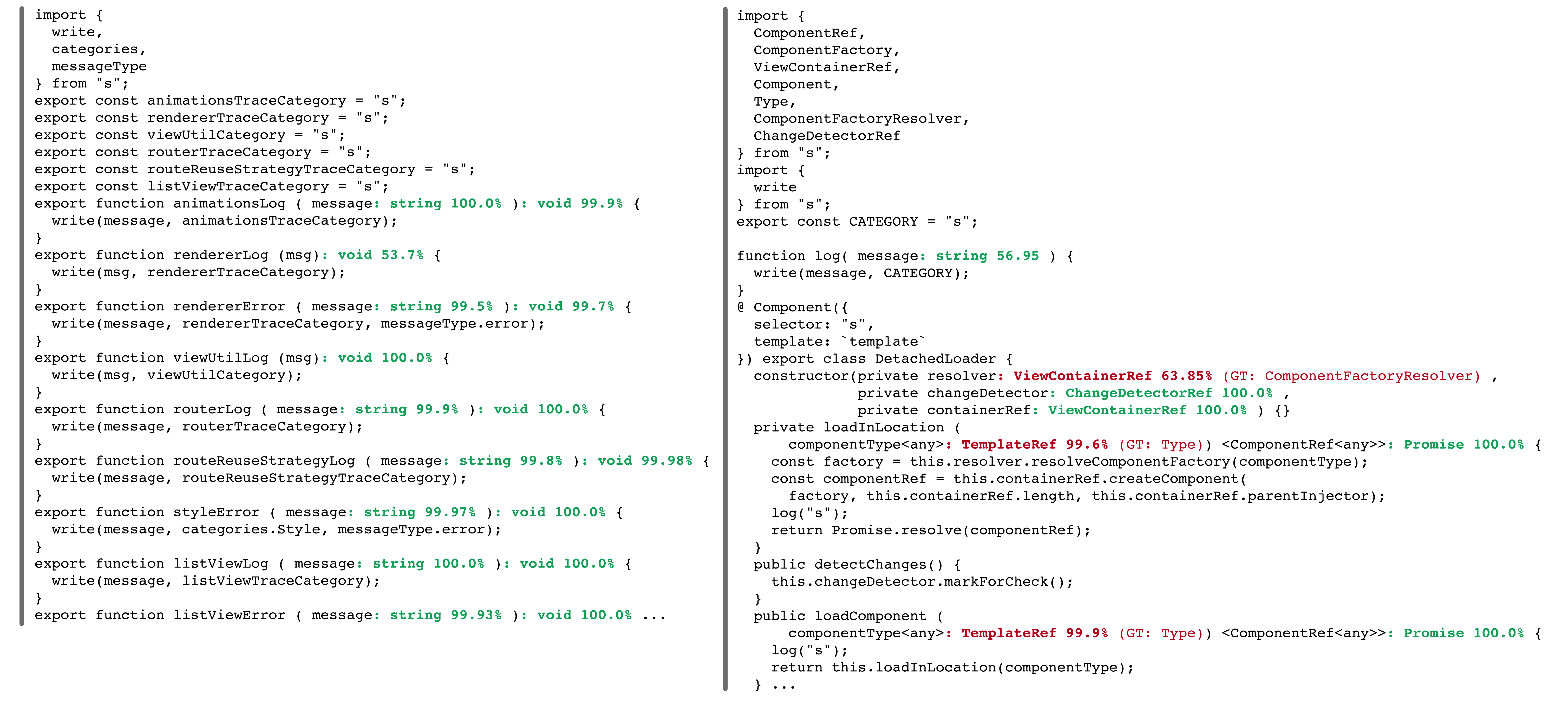}
    \caption{A variant of DeepTyper pre-trained with \ours{} generates type annotations for two held-out programs. The model consistently predicts correct function return types, and often correctly predicts project-specific variable types imported at the top of the file. Metrics are in the top row of Table~\ref{tab:mean_hidden_ablation} (not our best performing model).}
    \label{fig:qual_types}
\end{figure*}

In Table~\ref{tab:types}, contrastive pre-training outperforms all baseline learned methods.
\ours{} is applied in a drop-in fashion to each of the baselines. Pre-training with our contrastive objective and data augmentations yields absolute accuracy improvements of +1.2\%, +6.3\%, +2.3\% top-1 and +1.8\%, +5.7\%, +2.8\% top-5 over the Transformer, RoBERTa, and DeepTyper, respectively.

The RoBERTa baseline may perform poorly since the MLM objective focuses on token reconstruction that is overly sensitive to local syntactic structure, or because sufficient fine-tuning data is available, described as weight ``ossification'' by \citet{hernandez2021scaling}. To combine the approaches, we minimized our loss in addition to MLM as a hybrid local-global objective to pre-training a Transformer, improving accuracy by +6.31\% over the RoBERTa Transformer.

We also evaluate the recent GPT-3 Codex model by OpenAI~\cite{openai_codex} using their API. We benchmark the 175B parameter DaVinci model in both a zero-shot as well as a few-shot prompting setup. Although the Codex model was trained over TypeScript programs, it performs poorly as it achieves an accuracy of 26.6\% in the zero-shot setup and 30.6\% in the few-shot setup. We only evaluate Top-1 accuracy for GPT-3 models as GPT-3 does not reliably output confidence scores.

Learning outperforms static analysis by a large margin. Overall, our best model has +8.9\% higher top-1 accuracy than the built-in TypeScript CheckJS type inference system, showing the promise of learned code analysis. Surfacing multiple candidate types can also be useful to users, while CheckJS only has a single prediction.

Fig.~\ref{fig:qual_types} shows two files from held-out repositories.
For the first, our model consistently predicts the correct return and parameter types.
The model correctly predicts that the variable \codesnip{message} is a string, even though its type is ambiguous without access to the imported \codesnip{write} method signature. For the second, \ours{} predicts 4 of 8 types correctly including \codesnip{ViewContainerRef} and \codesnip{ChangeDetectorRef} from the AngularJS library. %

\begin{table}
\centering
\setlength\tabcolsep{4.2pt}
\centering
\begin{tabular}{lccc} \hline
    \textbf{Method} & \textbf{Precision} & \textbf{Recall} & \textbf{F1} \\
    \hline
    code2vec & 10.78\% & 8.24\% & 9.34\% \\
    code2seq & 12.17\% & 7.65\% & 9.39\% \\
    RoBERTa MLM & 15.13\% & 11.47\% & 12.45\% \\
    Transformer & 18.11\% & 15.78\% & 16.86\% \\
    \rowcolor{Gray}
    ~~+ \ours{} & 20.34\% & 14.96\% & \textbf{17.24\%}  \\  %
    \hline
\end{tabular}
\caption{Results for different settings of \textbf{code summarization}: supervised training with 81\thou{} functions, masked language model pre-training, training from scratch and contrastive pre-training with fine-tuning.}
\label{tab:summarization}
\end{table}

\subsection{Extreme Code Summarization}
\label{sec:experiments_code_summarization}
The extreme code summarization task asks a model to predict the name of a method given its body~\citep{allamanis2016convolutional}. These names often summarize the method, such as \codesnip{reverseString(...)}. Summarization models could help programmers interpret poorly documented code. We create a JavaScript summarization dataset using the 81,487 labeled methods in the CodeSearchNet dataset. The name is masked in the method declaration. A sequence-to-sequence model with an autoregressive decoder is trained to maximize log likelihood of the ground-truth name, a form of abstractive summarization. All models overfit, so we stop early according to validation loss. As proposed by~\citet{allamanis2016convolutional}, we evaluate model predictions by precision, recall and F1 scores over the set of method name tokens.

\begin{figure}
\begin{minipage}{0.5\linewidth}
\begin{footnotesize}
\begin{lstlisting}[language=JavaScript,basicstyle=\ttfamily\tiny]
function x(url, callback, error) {
  var img = new Image();
  img.src = url;
  if(img.complete){
    return callback(img);
  }
  img.onload = function(){
    img.onload = null;
    callback(img);
  };
  img.onerror = function(e){
    img.onerror = null;
    error(e);
  };
}	
\end{lstlisting}
\end{footnotesize}
\end{minipage}\hfill%
\begin{minipage}{0.5\linewidth}
\begin{small}
\textit{Ground truth}: \texttt{loadImage}\\
\textit{Prediction:} {\color{ForestGreen} \texttt{loadImage} }\\

Top predictions:
{\tiny
\begin{enumerate}
    \item \texttt{getImageItem}
    \item \texttt{createImage}
    \item \texttt{loadImageForBreakpoint}
    \item \texttt{getImageSrcCSS}
\end{enumerate}
}
\end{small}
\end{minipage}
\caption{A held-out JavaScript program from CodeSearchNet and method names generated by a Transformer pre-trained with \ours{}. The correct method name is predicted as the most likely decoding.}
\label{fig:qual_method_name}
\end{figure}

Table~\ref{tab:summarization} shows results in four settings: (1) supervised training using baseline tree-structured architectures that analyze the AST (code2vec, code2seq), (2) pre-training on all 1.8\million{} programs using MLM followed by fine-tuning on the labeled programs (RoBERTa), (3) training a Transformer from scratch and (4) contrastive pre-training followed by fine-tuning with augmentations.

Contrastive pre-training outperforms code2seq by +8.2\% test precision, +7.3\% recall, and +7.9\% F1 score.
\ours{} outperforms self-supervised pre-training with RoBERTa by +4.8\% F1.
\ours{} also achieves slightly higher performance than the Transformer learned from scratch. While this improvement is smaller, code summarization challenging as identifier names are not consistent between programmers.

Figure~\ref{fig:qual_method_name} shows a qualitative example of predictions for the code summarization task. The JavaScript method is not seen during training. A Transformer pre-trained with \ours{} predicts the correct method name through beam search. The next four predictions are reasonable, capturing that the method processes an image. The 2nd and 3rd most likely decodings, \codesnip{getImageItem} and \codesnip{createImage}, use \codesnip{get} and \codesnip{create} as synonyms for \codesnip{load}, though the final two unlikely decodings include terms not in the method body.

\begin{table}
\centering
\centering
\begin{tabular}{lc} \hline
    \cellcolor{white}\textbf{Code summarization model} & \textbf{F1} \\ \hline
    Transformer (Table~\ref{tab:summarization})              & \textbf{16.86}\\
    ~~~~+ augmentations                                 & 15.65\\ \hline \vspace{-2mm} \\ \hline
    \textbf{Type inference model} & \textbf{Acc@1} \\ \hline
    Transformer (Table~\ref{tab:types})     & \textbf{45.66}\\
    ~~~~+ augmentations                       & 44.14\\ \hline
    DeepTyper (Table~\ref{tab:types})       & \textbf{51.73}\\
    ~~~~+ augmentations              & 50.33\\
     \hline
\end{tabular}
\caption{Compiler data augmentations degrade performance when training supervised models \textit{from scratch}.}
\label{tab:supervised_aug}
\end{table}

\subsection{Understanding augmentation importance}
\label{sec:experiments_augmentation}
We analyze the effect of augmentations on supervised learning and on pre-training.

\paragraph{Supervised learning with augmentations} As a baseline, we re-train models from scratch with compiler transforms during \textit{supervised learning} rather than pre-training. Data augmentation artificially expands labeled training sets. For sequence-to-sequence summarization, we apply a variety of augmentations (LS, SW, VR, DCI). These all preserve the method name. For type inference, labels are aligned to input tokens, so they must be realigned after transformation. We only apply token-level transforms (LS, SW) as we can track labels.

Table~\ref{tab:supervised_aug} shows results. Compiler-based data augmentations degrade supervised models, perhaps by creating a training distribution not reflective of evaluation programs. However, as shown in \S\ref{sec:experiments_code_clone}--\ref{sec:experiments_code_summarization}, augmenting during \ours{} pre-training yields a more accurate model. Our contrastive learning framework also allows learning over large numbers of unlabeled programs that supervised learning alone cannot leverage. The ablation indicates that augmentations do not suffice, and contrastive learning is important.

\paragraph{Ablating pre-training augmentations} Some data augmentations could be more valuable than others. Empirically, pre-training converges faster with a smaller set of augmentations at the same batch size since the positives are syntactically more similar, but this hurts downstream performance. 
Table~\ref{tab:pretrain_aug_ablation} shows that type inference accuracy degrades when different groups of augmentations are removed. Semantics-preserving code compression passes that require code analysis are the most important, improving top-1 accuracy by 1.95\% when included. Line subsampling serves as a regularizer, but changes program semantics. LS is relatively less important, but does help accuracy. Identifier modifications preserve semantics, but change useful naming information.

\begin{table}
\setlength\tabcolsep{2pt}
\centering
\resizebox{\linewidth}{!}{
\begin{tabular}{lHcc} \hline
    \textbf{Pre-training augmentations} & \textbf{Fine-tuning} & \textbf{Acc@1} & \textbf{Acc@5}\\
    \hline
    All augmentations (Table~\ref{tab:types})                   & - & \textbf{52.65\%} & \textbf{84.60\%}\\
    ~~w/o identifier modification {\small(-VR, -IM)}         & - & 51.94\% & 84.43\%\\  %
    ~~w/o line subsampling {\small(-LS)}                   & - & 51.05\% & 81.63\%\\  %
    ~~w/o code compression {\small(-T,C,DCE,CF)} & - & 50.69\% & 81.95\%\\  %
    \hline 
\end{tabular}
}
\captionof{table}{Ablating compiler transformations used during contrastive pre-training. The DeepTyper BiLSTM is pre-trained with constrastive learning for 20\thou{} steps, then fine-tuned for type inference. Augmentations are only used during pre-training. Each transformation contributes to accuracy.}
\label{tab:pretrain_aug_ablation}
\end{table}

\section{Conclusion}
Large-scale code repositories like GitHub are a powerful resource for learning machine-aided programming tools. However, most current code representation learning approaches need labels, and popular label-free self-supervised methods like RoBERTa are not robust to adversarial inputs.
Instead of reconstructing tokens like BERT, learning \textit{what code says}, we learn \textit{what code does}. We propose \ours{}, a contrastive self-supervised algorithm that learns representations invariant to transformations via compiler-based data augmentations. In experiments, \ours{} learns effective representations of functionality, and is robust to adversarial code edits. We find that \ours{} significantly improves performance on three downstream JavaScript code understanding tasks.

\section*{Acknowledgments}
 We thank Lisa Dunlap, Jonathan Ho, Koushik Sen, Rishabh Singh, Aravind Srinivas, Daniel Rothchild, and Justin Wong for helpful feedback.
In addition to NSF CISE Expeditions Award CCF-1730628, the NSF GRFP under Grant No. DGE-1752814, and ONR PECASE N000141612723, this research is supported by gifts from Amazon Web Services, Ant Financial, Ericsson, Facebook, Futurewei, Google, Intel, Microsoft, NVIDIA, Scotiabank, Splunk and VMware.

\bibliography{main}
\bibliographystyle{acl_natbib}
\clearpage

\appendix
\section*{Appendices}
\section{Program transformation details}
\label{sec:appendix:program_transformations}
We use the Babel compiler infrastructure \citep{babel_github} and the \codesnip{terser} JavaScript library for AST-based program transformations. We perform variable renaming and dead code insertion (variable declaration insertion) using custom Babel transforms, subword regularization with \codesnip{sentencepiece} Python tokenization library, line subsampling using JavaScript string manipulation primatives and other transformations with \codesnip{terser}. Terser has two high-level transformation modes, mangling and compression, each with finer grained controls such as formatting, comment and log removal, and dead code elimination. We show an example merge sort with variants in Figure~\ref{fig:merge_sort_example}.

\textbf{Reformatting, beautification, compression (R, B, C):}
Personal coding conventions do not affect the semantics of code; auto-formatting normalizes according to a style convention.

\textbf{Dead-code elimination (DCE):}
In this pass, all unused code with no side effects are removed. Various statements can be inlined or removed as stale or unneeded functionality.

\textbf{Type upconversion (T):}
In JavaScript, some types are polymorphic \& can be converted between each other. As an example, booleans can be represented as \codesnip{true} or as \codesnip{1}.

\textbf{Constant folding (CF):}
During constant folding, all expressions that can be pre-computed at compilation time can be inlined. For example, the expression \codesnip{(2 + 3) * 4} is replaced with \codesnip{20}.

\textbf{Variable renaming, identifier mangling (VR, IM):}
Arguments can be renamed with random word sequences and identifiers can be replaced with short tokens to make the model robust to naming choices. Program behavior is preserved despite obfuscation.

\textbf{Dead-code insertion (DCI):}
Commonly used no-ops such as comments and logging are inserted.

\textbf{Subword regularization (SW):}
From~\citet{kudo2018subword}, text is tokenized in several different ways, with a single word (\codesnip{\_function}) or subtokens (\codesnip{\_func tion}).

\textbf{Line subsampling (LS):}
We randomly sample ($p = 0.9$) lines from a method body. While not semantics-preserving, line subsampling serves as a regularizer.

\begin{figure}[th!]
\centering
\begin{tcolorbox}[enhanced,size=small,colback=black!5!white,colframe=RoyalBlue,flip title={interior hidden},title={Original merge sort program}]
\begin{lstlisting}[language=JavaScript]
// Split the array into halves and merge them recursively
function mergeSort (arr) {
  if (arr.length === 1) {
    // return once we hit an array with a single item
    return arr
  }
  const middle = Math.floor(arr.length / 2)
  // get the middle item of the array rounded down
  const left = arr.slice(0, middle)
  // items on the left side
  const right = arr.slice(middle)
  // items on the right side
  return merge(
    mergeSort(left),
    mergeSort(right)
  )
}
\end{lstlisting}
\end{tcolorbox}
\vspace{2pt}

\begin{tcolorbox}[enhanced,size=small,colback=black!5!white,colframe=RoyalBlue,flip title={interior hidden},title={Variable renaming, comment removal, reformatting}]
\begin{tiny}
\begin{lstlisting}[language=JavaScript]
function mergeSort(e) {
  if (e.length === 1) {
    return e;
  }
  const t = Math.floor(e.length / 2);
  const l = e.slice(0, t);
  const n = e.slice(t);
  return merge(mergeSort(l), mergeSort(n));
}
\end{lstlisting}
\end{tiny}
\end{tcolorbox}
\vspace{2pt}

\begin{tcolorbox}[enhanced,size=small,colback=black!5!white,colframe=RoyalBlue,flip title={interior hidden},title={Combining variable declarations, inlining conditional}]
\begin{tiny}
\begin{lstlisting}[language=JavaScript]
function mergeSort(e) {
  if (1 === e.length) return e;
  const t = Math.floor(e.length / 2), r = e.slice(0, t), n = e.slice(t);
  return merge(mergeSort(r), mergeSort(n));
}
\end{lstlisting}
\end{tiny}
\end{tcolorbox}
\caption{Given a JavaScript code snippet implementing the merge sort algorithm, we apply semantics-preserving transformations to produce functionally-equivalent yet textually distinct code sequences. Variable renaming and identifier mangling passes change variable names. Compression passes eliminate unnecessary characters such as redundant variable declarations and brackets.}
\label{fig:merge_sort_example}
\end{figure}

\begin{figure*}
  \centering
  \includegraphics[width=0.75\textwidth]{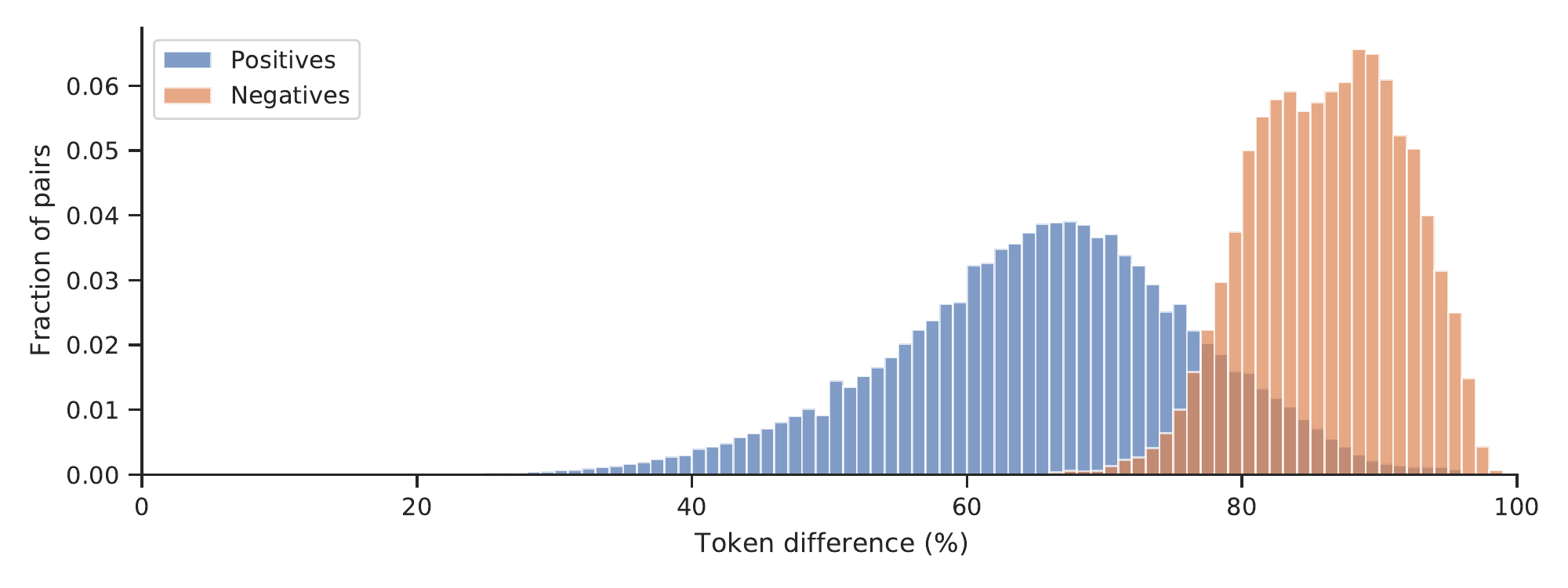}
  \caption{Histogram of pairwise token dissimilarity for contrastive positives (transformed variants of the same method) and negatives (transformed variants of different methods). Code transformations produce positives with dissimilar token sequences.}
  \label{fig:tokendistance}
\end{figure*}

\section{How similar are transformed programs?}
\label{sec:appendix:dissimilarity}
To understand the diversity created by program transformations, we compute the Levenshtein minimum edit distance between positive pairs in the precomputed pre-training dataset, \textit{i.e.} transformed variants of the same source method. For comparison, we also compute the edit distance between negative pairs: transformed variants of different programs.

The edit distance $D(x_q, x_k)$ computes the minimum number of token insertions, deletions or substitutions needed to transform the tokenized query progrm $x_q$ into the key program $x_k$.
 To normalize by sequence length $|\cdot|$, let \begin{equation}
    \text{dissimilarity}_{D}(x_q, x_k) = \frac{D(x_q, x_k)}{\max(|x_q|, |x_k|)}
    \label{eq:dissimilarity}
\end{equation}

Dissimilarity ranges from 0\% for programs with the same sequence of tokens, to 100\% for programs without any shared tokens.
Note that whitespace transformations do not affect the metric because the tokenizer collapses repeated whitespace. For the positives, we estimate dissimilarity by sampling one pair per source program in the CodeSearchNet dataset (1.6M source programs with at least one pair). We sample the same number of negative pairs.

Fig.~\ref{fig:tokendistance} shows a histogram of token dissimilarity. Positive pairs have $65\%$ mean dissimilarity, while negatives have $86\%$. Negatives are more dissimilar on average as source sequences could have different lengths, idioms and functionality. Still, the transformations generated quite different positive sequences, with less than half of their tokens shared. The 25th, median and 75th percentile dissimilarity is 59\%, 66\% and 73\% for positives, and 82\%, 87\% and 90\% for negatives.

\section{Experimental setup}
\textbf{Architectures}~~~~
The Transformer encoder has 6 layers (23M parameters) in all experiments. For code summarization experiments, we add 4 decoder layers with causal masking to generate the natural language summary.
We leverage the default positional embedding function ($\sin$, $\cos$) as used in the original Transformer architecture. The network originally proposed in DeepTyper \citep{hellendoorn2018deep} had 11M parameters with a 300 dimensional hidden state. We increase the hidden state size to 512 to increase model capacity, so our BiLSTM for type prediction has 17.5M parameters.
During fine-tuning, across all experiments, we optimize parameters using Adam with linear learning rate warmup and decay. For the Transformer, the learning rate is linearly increased for 5,000 steps from 0 to a maximum of $10^{-4}$. For the bidirectional LSTM, the learning rate is increased for between 2,500 and 10,000 steps to a maximum of $10^{-3}$. Type inference hyperparameters are selected by validation top-1 accuracy.

\textbf{\ours{} pre-training}~~~~The InfoNCE objective is minimized with temperature $t=0.07$ following~\citet{he2019momentum}. Also following \citet{he2019momentum}, the key encoder's parameters are computed with the momentum update equation $\theta_{k} \leftarrow m\theta_{k} + (1-m)\theta_{q}$, equivalent to an EMA of the query encoder parameters $\theta_{q}$. To pretrain a Transformer using the \ours{} objective, we first embed each token in the program using the Transformer. However, the InfoNCE objective is defined in terms of a single embedding for the full program. The \ours{} Transformer is pre-trained with a batch size of 96. Our model averages the 512-dimensional token embeddings across the sequence, then applies a two-layer MLP with 512 hidden units and a ReLU activation to extract a 128-dimensional embedding for the loss.

The DeepTyper bidirectional LSTM architecture has two choices for extracting a global program representation. We aggregate a 1024-dimensional representation of the program by concatenating its four terminal hidden states (from two sequence processing directions and two stacked LSTM layers), then apply the same MLP architecture as before to extract a 128-dimensional representation. Alternatively, we can average the hidden state concatenated from each direction across the tokens in the sequence before applying the MLP head. We refer to the hidden-state configuration as a global representation and the sequence averaging configuration as a local representation in Tab.~\ref{tab:mean_hidden_ablation}. We pre-train the BiLSTM with large batch size of 512 and apply weight decay.

\textbf{Code clone detection on HackerRank programs}~~~~Figure~\ref{fig:code_clone_example} shows two programs sampled from the HackerRank clone detection dataset. These programs successfully solve the same problem, so they are clones. We report metrics that treat code clone detection as a binary classification task given a pair of programs. 2065 pairs of programs solving the same HackerRank problem and 2065 pairs of programs solving different problems are sampled to construct an evaluation dataset. We use the area under the Receiver Operating Characteristic (AUROC) metric and Average Precision (AP) metrics. The standard error of the AUROC is reported according to the Wilcoxon statistic~\cite{10.5555/1089508.1089530}. Average Precision is the area under the Precision-Recall curve. AUROC and AP are both computed using the \codesnip{scikit-learn} library~\cite{scikit-learn}.

A Transformer predicts contextual embeddings of each token in a program, but our thresholded cosine similiarity classifier requires fixed length embeddings of whole programs. To determine if two programs that may differ in length are clones, we pool the token representations across the sequence. We evaluated both mean pooling and max pooling the representation. For the hybrid model pre-trained with both RoBERTa (MLM) and contrastive objectives, mean pooling achieved the best AUROC and AP. For other models, max pooling performed the best.

\textbf{Type prediction}~~~~Following DeepTyper~\citep{hellendoorn2018deep}, our regenerated dataset for type prediction has
187 training projects with 15,570 TypeScript files, totaling 6,902,642 tokens. We tune hyperparameters on a validation set of 23 distinct projects with 1,803 files and 490,335 tokens, and evaluate on a held-out test set of 24 projects with 2,206 files and 958,821. The training set is smaller than originally used in DeepTyper as several projects were made private or deleted from GitHub before May 2020 when we downloaded the data, but we used the same commit hashes for available projects so our splits are a subset of the original. We have released the data with our open-source code to facilitate further work on a stable benchmark as more repositories are deleted over time. We perform early stopping to select the number of training epochs. We train each model for 100 epochs and select the checkpoint with the minimum accuracy@1 metric (all types, including \codesnip{any}) on the validation set. Except for the model learned from scratch, the Transformer architectures are pre-trained for 240\thou{} steps. Models with the DeepTyper architecture converge faster on the pre-training tasks and are pre-trained for 20\thou{} iterations (unless otherwise noted).

\begin{figure}[t!]
    \centering
    \subfloat[Character length per code sample]{{\includegraphics[width=0.3\textwidth]{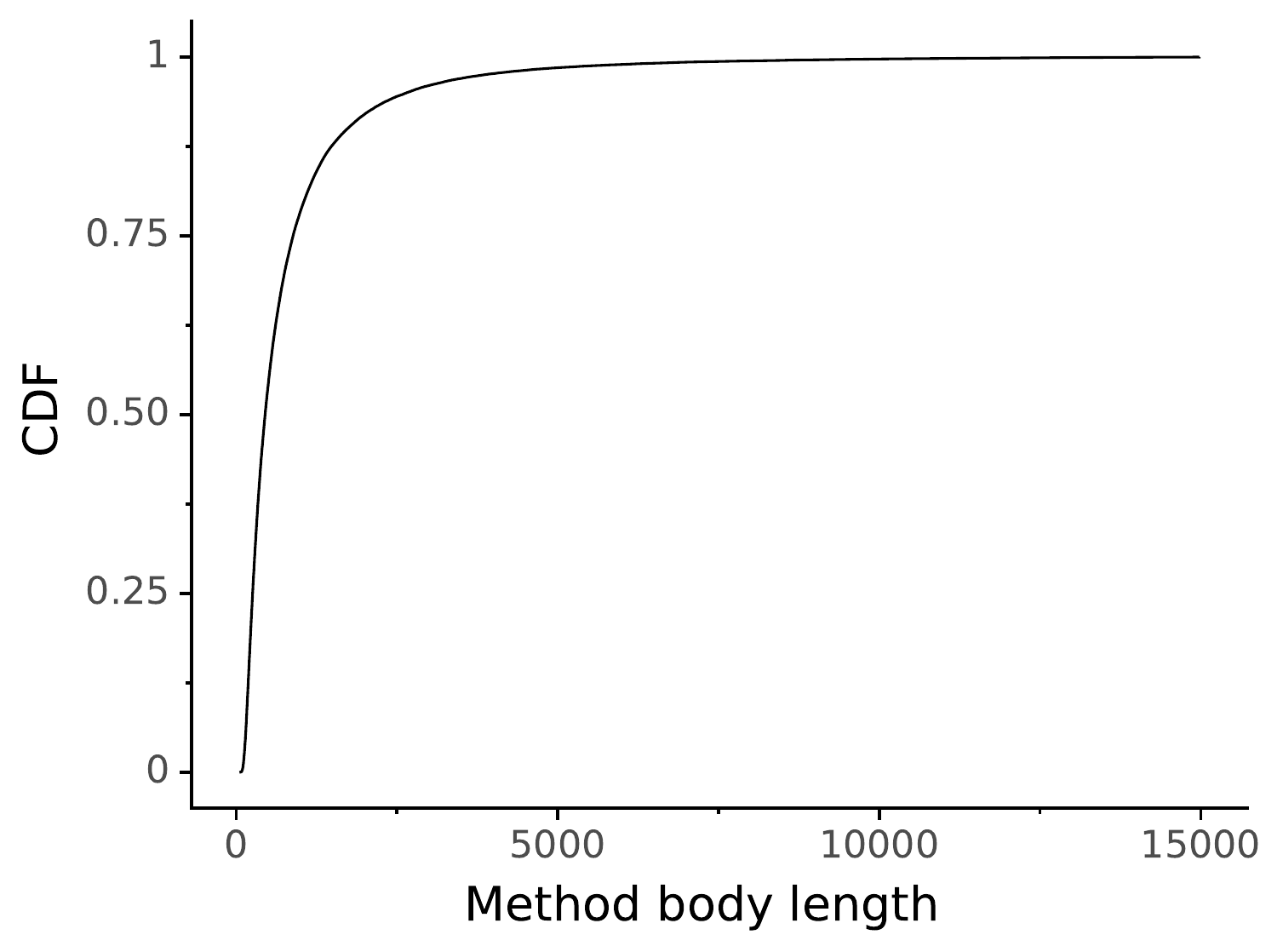} }}%
    \qquad
    \subfloat[Character length per method name]{{\includegraphics[width=0.3\textwidth]{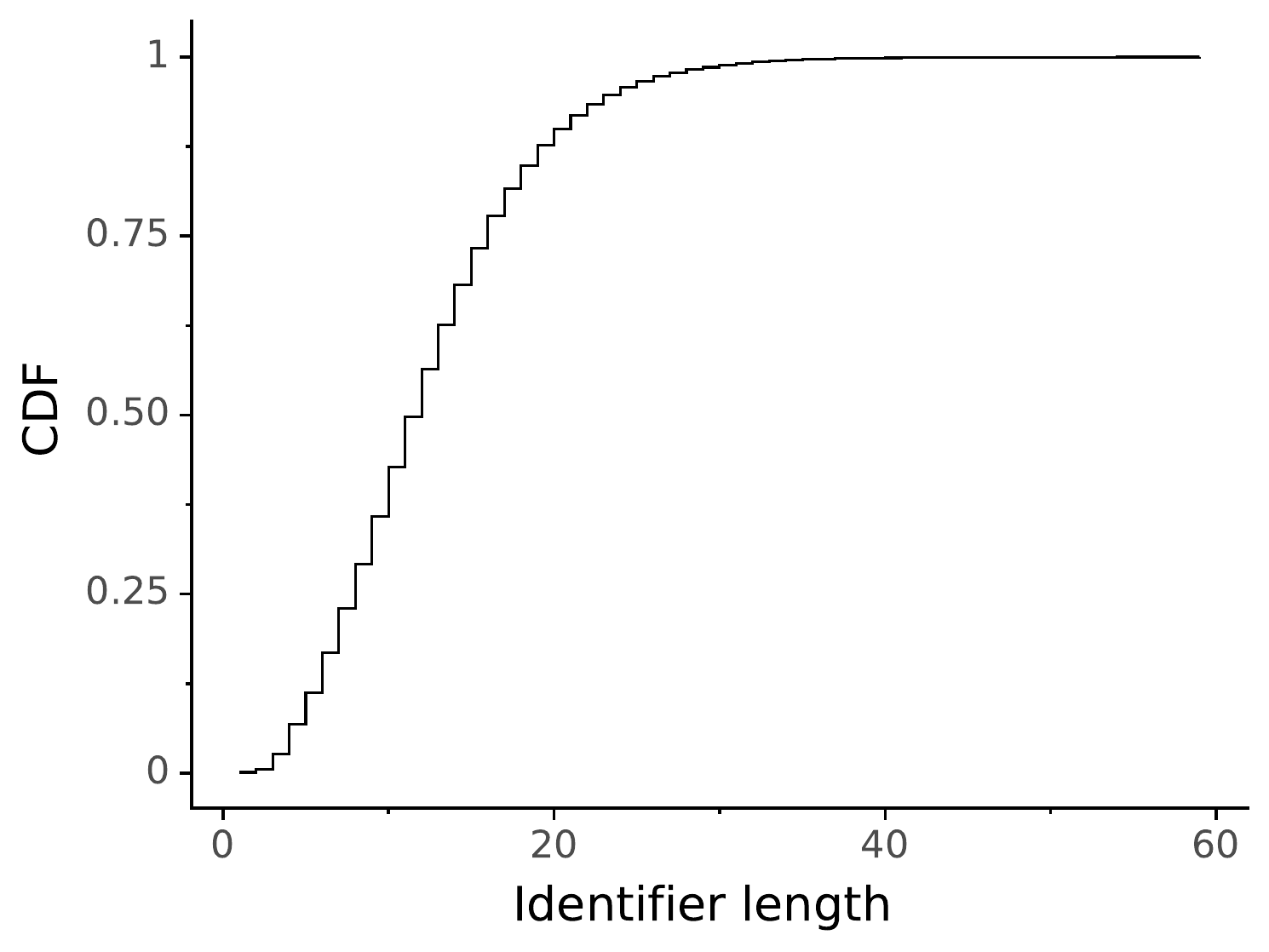} }}%
    \caption{CodeSearchNet code summarization dataset statistics: (a) The majority of code sequences are under 2000 characters, but there is long tail of programs that span up to 15000 characters long, (b) JavaScript method names are relatively short compared to languages like C$^\sharp$ and Java.}
    \label{fig:dataset_statistics_identifier}
\end{figure}

\textbf{Extreme code summarization by method name prediction}~~~~We train method prediction models using the labeled subset of CodeSearchNet. Neither method names nor docstrings are provided as input to the model: the docstring is deleted, and the method name is replaced with the token `\codesnip{x}'. Thus, the task is to predict the method name using the method body and comments alone.

To decode method names from all models except the code2vec and code2seq baselines which implement their own decoding procedures, we use a beam search with a beam of size $5$ and a maximum target sequence length of $20$ subword tokens. We detail the cumulative distribution of program lengths in Figure~\ref{fig:dataset_statistics_identifier}. The \ours{} summarization Transformer only needed to be pre-trained for 20\thou{} iterations, with substantially faster convergence than RoBERTa (240\thou{} iterations). During fine-tuning, we apply the LS,SW,VR,DCI augmentations to \ours{}.

\section{Baselines}
\label{sec:appendix:baseline_js}
Baselines for code summarization and type prediction trained their models on an inconsistent set of programming languages and datasets. In order to normalize the effect of datasets, we selected several diverse state-of-the-art baselines and reimplemented them on the JavaScript dataset.

\textbf{AST-based models}~~~~The authors of code2vec~\citep{alon2019code2vec} and code2seq~\citep{alon2018code2seq}, AST-based code understanding models, made both data and code available, but train their model on the Java programming language. In order to extend the results in their paper to JavaScript for comparison with our approach, we generated an AST path dataset for the CodeSearchNet dataset.
The sensitivity of path-mining embeddings to different datasets is documented in prior work, so published F1 scores are not directly comparable; F1 scores for code2vec~\citep{alon2019code2vec} vary between 19 \citep{alon2018code2seq} and 43 \citep{alon2019code2vec} depending on the dataset used. Therefore, we use the same dataset generation code as the authors for fair comparison. We first parse the source functions using the Babel compiler infrastructure. Using the original code on these ASTs, up to 300 token-to-token (leaf-to-leaf) paths are extracted from each function's AST as a precomputed dataset. Then, we generate a token and AST node vocabulary using the same author-provided code, and train the models for 20 epochs, using early stopping for code2seq. We observed that code2vec overfits after 20 epochs, and longer training was not beneficial.

\textbf{DeepTyper}~\citep{hellendoorn2018deep}~~~~DeepTyper uses a two layer GRU with a projection over possible classes, with an embedding size of 300 and hidden dimension of 650. However, we found improved performance by replacing the GRU with a bidirectional LSTM (BiLSTM). We normalize the LSTM parameter count to match our model, and therefore use a hidden dimension size of 512. We also use subword tokenization rather than space delimited tokens according to \citet{kudo2018subword}, as subwords are a key part of state-of-the-art models for NLP~\citep{sennrich2015neural}.

\textbf{RoBERTa}~~~~We pre-trained an encoder using RoBERTa's masked language modeling loss on our augmented version of CodeSearchNet, the same data used to pre-train \ours{}. This model is then fine-tuned on downstream datasets. Unlike the original BERT paper which cuBERT~\citep{cuBERT} is based on, hyperparameters from RoBERTa have been found to produce better results during pre-training. RoBERTa pre-trains using a masked language modeling (MLM) objective, where 15\% of tokens in a sentence are masked or replaced and are reconstructed by the model. We did not use the BERT Next Sentence Prediction (NSP) loss which RoBERTa finds to be unnecessary. We normalize baseline parameter count by reducing the number of Transformer layers from 24 to 6 for a total of 23M parameters.

\section{Additional results and ablations}
\label{sec:appendix_ablations}

\begin{figure}[t]
    \centering
    \includegraphics[width=\linewidth]{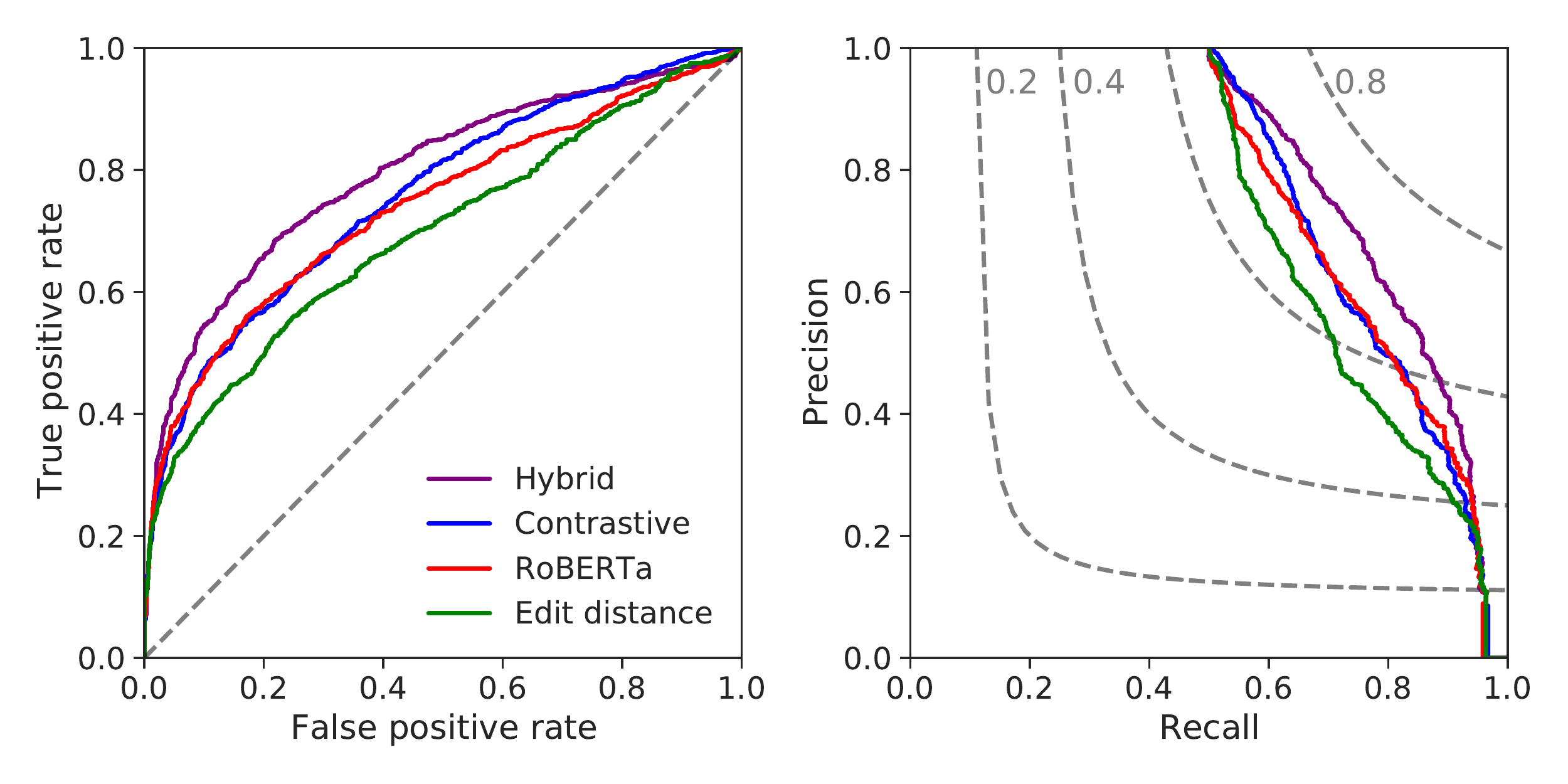}
    \caption{Receiver Operating Characteristic (ROC) and Precision-Recall (PR) curves for non-adversarial classifiers on the code clone detection task. Equal F1 score curves are shown on right.}
    \label{fig:codeclone_curves}
\end{figure}

\begin{figure}[t]
    \centering
    \includegraphics[width=\linewidth]{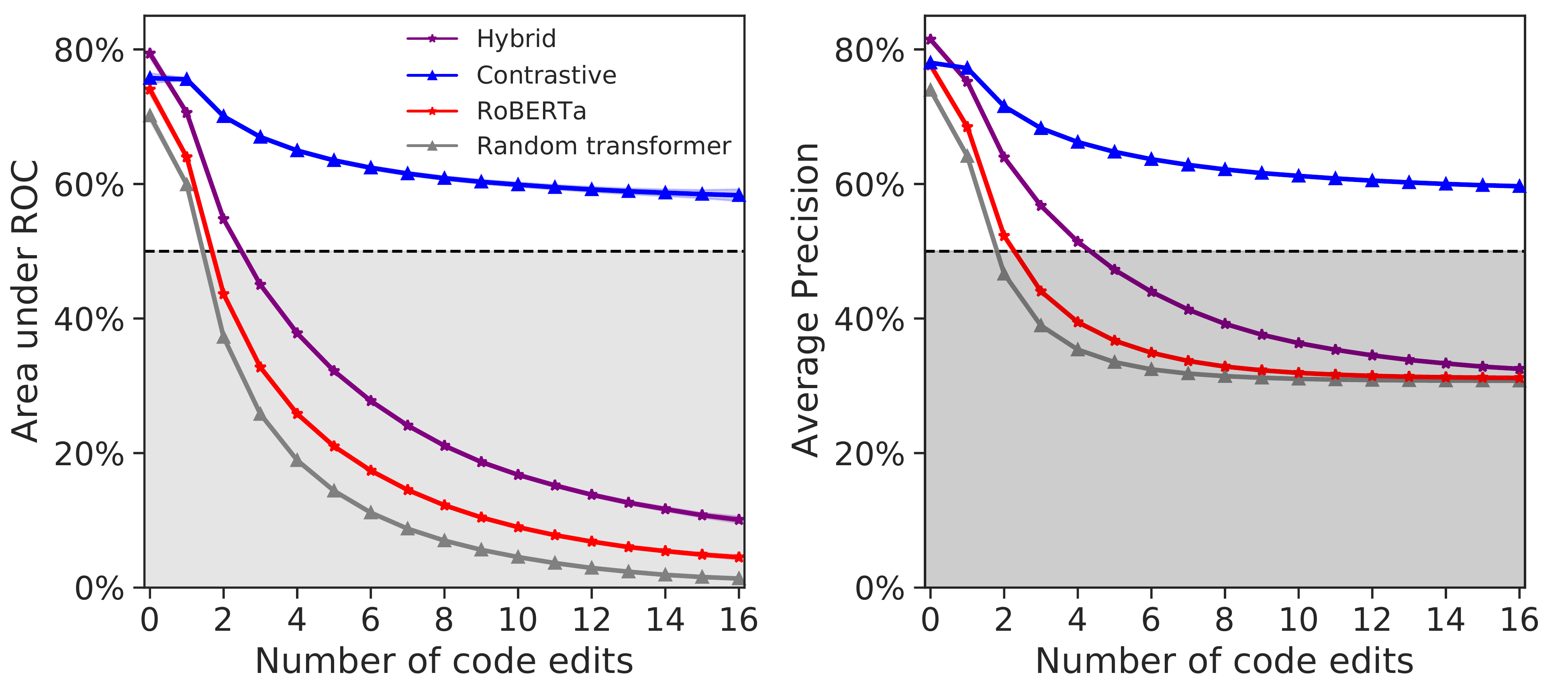}
    \caption{Adversarial AUROC and Average Precision for four models on the code clone detection task: a randomly initialized transformer, and transformers pre-trained on code with the RoBERTa MLM objective, our contrastive objective, or both. Representations learned by the contrastive model transfer robustly.}
    \label{fig:codeclone_auroc_ap}
\end{figure}

\textbf{Code clone detection ROC, PR curves}~~~~Fig.~\ref{fig:codeclone_curves} plots true postive rate vs false positive rate and precision vs recall for different zero-shot classifiers on the code clone detection downstream tasks. These classifiers threshold a similarity score given by token-level edit distance for the heuristic approach or cosine similarity for the neural network representations. The hybrid self-supervised model combining \ours{}'s contrastive objective and MLM achieves better tradeoffs than the other approaches. Fig.~\ref{fig:codeclone_auroc_ap} shows the AUROC and Average Precision of four Transformer models on the same task under adversarial transformations of one input program. Untrained models as well as models pre-trained with RoBERTa's MLM objective are not robust to these code transformations. However, the model pre-trained with \ours{} preserves much of its performance as the adversarial attack is strengthened.

\textbf{Which part of the model should be transferred?}~~~~
SimCLR \citep{chen2020simple} proposed using a small MLP head to reduce the dimensionality of the representation used in the InfoNCE loss during pre-training, and did not transfer the MLP to the downstream image-classification task. In contrast, we find it beneficial to transfer part of the contrastive MLP head to type inference, showing a $2\%$ improvement in top-5 accuracy over transferring the encoder only (Table~\ref{tab:mlp_transfer_ablation}). We believe the improvement stems from fine-tuning both the encoder and MLP which allows feature adaptation, while SimCLR trained a linear model on top of frozen features. We only transferred the MLP when contrasting the mean of token embeddings during pre-training, not the terminal hidden states, as the dimensionality of the MLP head differs. These representations are compared next.

\begin{table}
\caption{If local representations are learned, transferring part of the Contrastive MLP head improves type inference. The encoder is a 2-layer BiLSTM (d=512), with a 2-layer MLP head for both pre-training purposes and type inference. The mean hidden state representation is optimized for 10\thou{} iterations for the purposes of this ablation.}
\setlength\tabcolsep{3.5pt}
\label{tab:mlp_transfer_ablation}
\centering
\begin{tabular}{lccHH} \hline
    \textbf{Warm-started layers} & \textbf{Acc@1} & \textbf{Acc@5} & Acc@1 & Acc@5\\
    \hline
    BiLSTM & \textbf{49.32\%} & 80.03\% & 59.75\% & 77.98\% \\
    BiLSTM, 1 layer of MLP & 49.15\% & \textbf{82.58\%} & \textbf{60.88\%} & \textbf{79.41\%} \\
    \hline 
\end{tabular}
\end{table}

\begin{table*}
\caption{Contrasting global, sequence-level representations outperforms contrasting local representations. We compare using the terminal (global) hidden states of the DeepTyper BiLSTM and the mean pooled token-level (local) hidden states.}
\small
\label{tab:mean_hidden_ablation}
\centering
\begin{tabular}{clccHH} \hline
    \textbf{Representation} & \textbf{Optimization} & \textbf{Acc@1} & \textbf{Acc@5} & Acc@1 & Acc@5\\
    \hline
    \multirow{2}{*}{Global} & InfoNCE with terminal hidden state, 20\thou{} steps & \textbf{52.65\%} & \textbf{84.60\%} & \textbf{63.35\%} & \textbf{79.69\%} \\
    & InfoNCE with terminal hidden state, 10\thou{} steps & 51.70\% & 83.03\% & 62.16\% & 79.56\% \\ \hline
    Local & InfoNCE with mean token rep., 10\thou{} steps & 49.32\% & 80.03\% & 59.75\% & 77.98\% \\
    \hline 
\end{tabular}
\end{table*}

\begin{table*}
\caption{Training time and decoder depth ablation on the method name prediction task. Longer pre-training significantly improves downstream performance when a shallow, 1 layer decoder is used.} 
\label{tab:summarization_depth_ablation}
\footnotesize
\centering
\begin{tabular}{llcccc} \hline
    \multirow{2}{*}{\textbf{Decoder}} & \textbf{Pre-training} & \textbf{Supervision} & \multirow{2}{*}{\textbf{Precision}} & \multirow{2}{*}{\textbf{Recall}} & \multirow{2}{*}{\textbf{F1}}\\
    & (1.8M programs) & (81k programs) & \\
    \hline
    Transformer, 1 layer & MoCo, 10k steps & Original set & 11.91\% & 5.96\% & 7.49\%\\
    Transformer, 1 layer & MoCo, 45k steps & Original set & \textbf{17.71\%} & \textbf{12.57\%} & \textbf{13.79\%}\\
    Transformer, 4 layers & MoCo, 45k steps & Original set & \textbf{18.21\%} & \textbf{13.21\%} & \textbf{14.56\%} \\
     \hline 
\end{tabular}
\end{table*}

\textbf{Should we pre-train global or local representations?}~~~~We compare pre-training DeepTyper with two variants of \ours{}. We either use the mean of token hidden states across the program (averaging local features), or the terminal hidden states as input to the MLP used to extract the contrastive representation $q=f_q(x)$ (global features). Token-level features might capture more syntactic details, but averaging pooling ignores order. Table~\ref{tab:mean_hidden_ablation} shows the accuracy of a BiLSTM pre-trained with each strategy. Using the global features for pre-training yields significantly improved performance, +2.38\% acc@1 after 10\thou{} iterations of pre-training (not converged for the purposes of ablation). The global pre-training strategy achieves our best results.

\textbf{Do pre-trained encoders help more with shallow decoders?}~~~~
For the sequence-to-sequence code summarization task, \ours{} only pre-trains the encoder of the Transformer. In Table~\ref{tab:summarization_depth_ablation}, we ablate the depth of the decoder to understand how much shallow decoders benefit from contrastive pre-training of the encoder. Similar experiments were performed in a vision context by~\cite{erhan2010does}, where different numbers of layers of a classifier are pre-trained. After 45k pre-training steps, the 4-layer decoder achieves $0.50\%$ higher precision, $0.64\%$ higher recall and $0.77\%$ higher F1 score than the 1-layer model, so additional decoder depth is helpful for the downstream task. The 1-layer decoder model also benefits significantly from longer pre-training, with a $6.3\%$ increase in F1 from 10k to 45k iterations. This large of an improvement indicates that \ours{} could be more helpful for pre-training when the number of randomly initialized parameters at the start of fine-tuning is small. For larger decoders, more parameters must be optimized during-finetuning, and the value of pre-training is diminished.

\begin{figure}[t]
    \centering
    \includegraphics[width=0.5\linewidth]{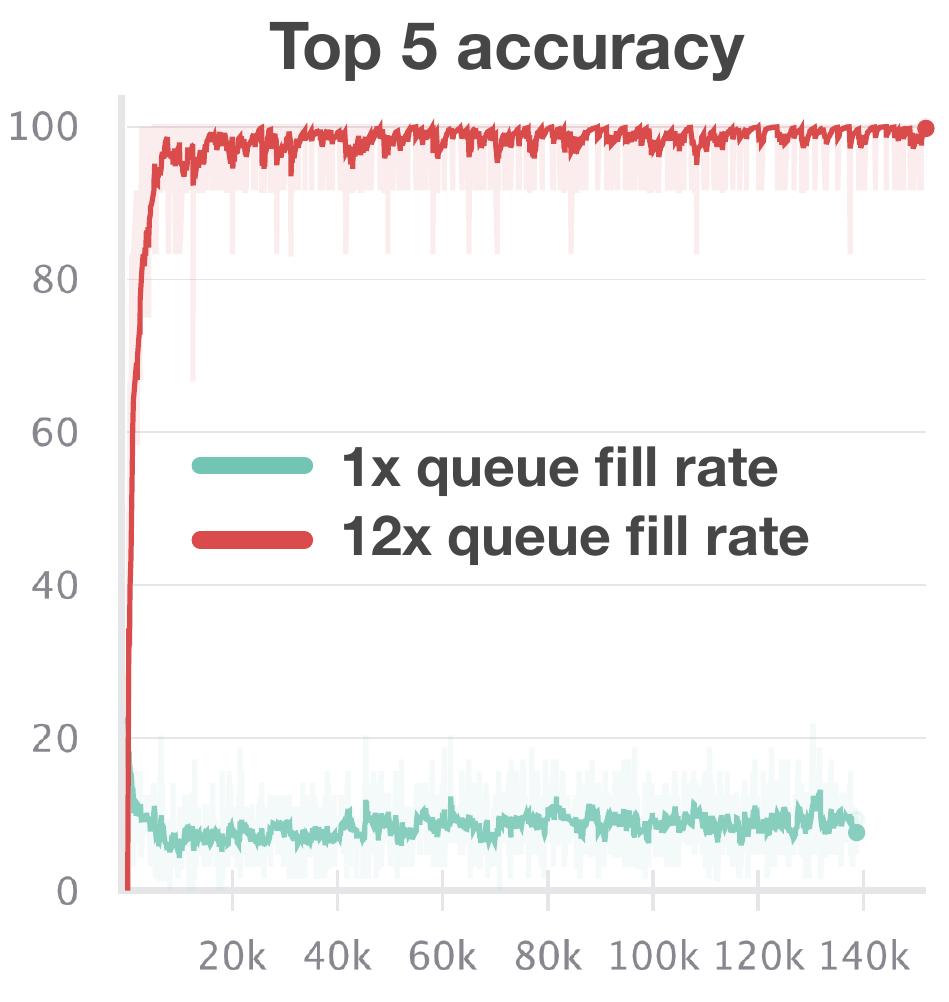}
    \caption{Pre-training quickly converges if negative programs in the queue are frequently changed.}
    \label{fig:constrastive_pretraining_accuracy}
\end{figure}

\textbf{Contrastive representation learning strategies}~~~~In Figure~\ref{fig:constrastive_pretraining_accuracy}, we compare two strategies of 
refreshing the MoCo queue of key embeddings (the dictionary of negative program representations assumed to be non-equivalent to the batch of positives). 
In the first strategy, we add 8 items out of the batch to the queue (1$\times$), while in the second we add 96 items (12$\times$). In addition, we use a larger queue (65k versus 125k keys) and a slightly larger batch size (64 versus 96).
We observe that for the baseline queue fill rate, the accuracy decreases for the first 8125 iterations as the queue fills. This decrease in accuracy is expected as the task becomes more difficult due to the increasing number of negatives during queue warmup. However, it is surprising that accuracy grows so slowly once the queue is filled. 
We suspect this is because the key encoder changes significantly over thousands of iterations: with a momentum term $m=0.999$, the original key encoder parameters are decayed by a factor of $2.9 \times 10^{-4}$ by the moving average. If the queue is rapidly refreshed, queue embeddings are predicted by recent key encoders, not old parameters. This also indicates that a large diversity of negative, non-equivalent programs are helpful for rapid convergence of \ours{} pre-training.

\begin{figure}
    \centering
    \includegraphics[width=\linewidth]{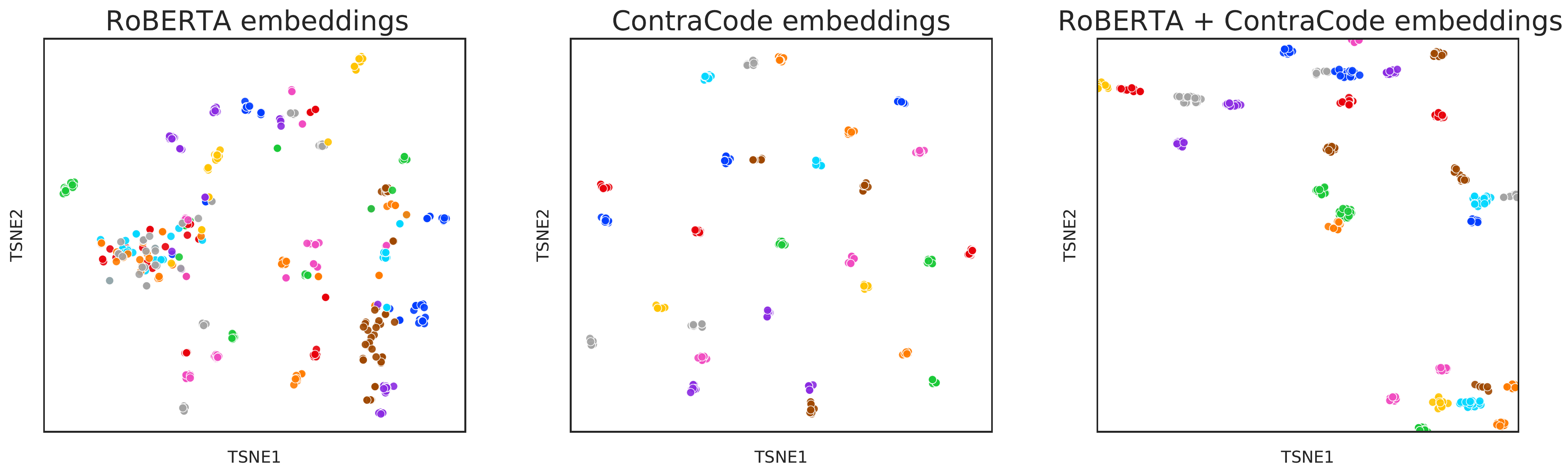}
    \caption{t-SNE~\citep{tsne2008} plot of mean pooled program representations learned with masked language modeling (RoBERTa), contrastive learning (\ours{}), and a hybrid loss (RoBERTa + \ours{}). Transformed variants of the same program share the same color. Note that colors may be similar across different programs.}%
    \label{fig:tsne}
\end{figure}

\paragraph{t-SNE visualization of representations} We qualitatively inspect the structure of the learned representation space by visualizing self-supervised representations of variants of 28 programs using t-SNE~\citep{tsne2008} in Figure~\ref{fig:tsne}. Representations of transformed variants of the same program are plotted with the same color. \ours{} (BiLSTM) clusters variants closely together. Indeed, contrastive learning learns representations that are invariant to a wide class of automated compiler-based transformations. In comparison, the representations learned by masked language modeling (RoBERTa) show more overlap between different programs, and variants do not cleanly cluster. With a hybrid loss combining masked language modeling and contrastive learning, representations of variants of the same program once again cluster.

\end{document}